\documentclass[10pt,twocolumn,letterpaper]{article}

\usepackage{cvpr}              
\ifdefined\checkmark
  \let\oldcheckmark\checkmark
  \undef\checkmark
\fi 
\usepackage{dingbat}
\usepackage[dvipsnames]{xcolor}

\definecolor{cvprblue}{rgb}{0.21,0.49,0.74}
\usepackage[pagebackref,breaklinks,colorlinks,citecolor=cvprblue]{hyperref}

\usepackage{makecell}
\usepackage{graphicx}
\usepackage{booktabs}
\usepackage{multirow} 
\usepackage{wrapfig} 
\usepackage{caption}

\usepackage{dingbat} 
\ifdefined\oldcheckmark
  \let\checkmark\oldcheckmark
\fi 

\def\paperID{4537} 
\def\confName{CVPR}
\def\confYear{2024}

\renewcommand{\paragraph}[1]{\vspace{2pt}\par\noindent\textbf{#1}~}

\title{Elucidating and Overcoming the Challenges of Label Noise\\in Supervised Contrastive Learning} 

\author{
    Zijun Long\thanks{Emails: z.long.2@research.gla.ac.uk, 2182951k@student.gla.ac.uk, 2658047Z@student.gla.ac.uk, richard.mccreadie@glasgow.ac.uk, gerardo.aragoncamarasa@glasgow.ac.uk, paul.henderson@glasgow.ac.uk. All authors are with the University of Glasgow.} \and
    George Killick\footnotemark[1] \and
    Lipeng Zhuang\footnotemark[1] \and
    Richard McCreadie\footnotemark[1] \and
    Gerardo Aragon-Camarasa\footnotemark[1] \and
    Paul Henderson\footnotemark[1]
}

\begin{document}
\maketitle

\begin{abstract}
Image classification datasets exhibit a non-negligible fraction of mislabeled examples, often due to human error when one class superficially resembles another.
This issue poses challenges in supervised contrastive learning (SCL), where the goal is to cluster together data points of the same class in the embedding space while distancing those of disparate classes. While such methods outperform those based on cross-entropy, they are not immune to labeling errors.
However, while the detrimental effects of noisy labels in supervised learning are well-researched, their influence on SCL remains largely unexplored.
Hence, we analyse the effect of label errors and examine how they disrupt the SCL algorithm's ability to distinguish between positive and negative sample pairs.
Our analysis reveals that human labeling errors manifest as easy positive samples in around 99\% of cases. 
We, therefore, propose D-SCL, a novel Debiased Supervised Contrastive Learning objective designed to mitigate the bias introduced by labeling errors. 
We demonstrate that D-SCL consistently outperforms state-of-the-art techniques for representation learning across diverse vision benchmarks, offering improved robustness to label errors.
\end{abstract}
    
\section{Introduction}
\label{sec:intro}

Contrastive methods achieve excellent performance on self-supervised learning \cite{RN89, npairloss, DBLP:journals/corr/abs-1810-06951}. They produce latent representations that excel at a multitude of downstream tasks, from image recognition and object detection to visual tracking and text matching \cite{suconlan,Tan2022DomainGF}. \textit{Supervised} contrastive learning (SCL) utilizes label information to improve representation learning, encouraging closer distances between same-class samples (positive pairs) and greater distances for different-class samples (negative pairs). SCL outperforms traditional methods for pre-training that employ a cross-entropy loss \citep{suconlan, li2022selective, RN81}.

However, the effectiveness of SCL depends on the quality of the labels used for supervision. Noisy labels introduce erroneous positive and negative pairings, compromising the integrity of learned representations \citep{li2022selective}.  
In practice, obtaining accurately annotated datasets is cost-prohibitive \cite{DBLP:conf/nips/NorthcuttAM21,welinder2010online}, and even widely-used datasets exhibit significant numbers of mislabeled images---e.g.~the ImageNet validation set has 5.83\% of images wrongly labeled \cite{DBLP:conf/nips/NorthcuttAM21}. Moreover, it is neither feasible to re-annotate all training samples across existing datasets nor to guarantee that new datasets are error-free \cite{DBLP:conf/nips/NorthcuttAM21,welinder2010online}. 
Thus, there is a critical need for methods that automatically mitigate the impact of annotation errors, enabling optimal use of existing datasets without requiring labor-intensive manual re-annotation.

The impact of noisy labels on supervised learning has been extensively researched \cite{DBLP:journals/tnn/SongKPSL23,DBLP:journals/corr/abs-2011-04406,DBLP:conf/iccS/DubelWN23}.
However, the extent and manner in which noisy labels influence SCL remains under-explored.
Hence, this paper investigates the effects of noisy labels on supervised contrastive learning, and develops an efficient but effective method to mitigate label noise, enabling more robust learning of representations.

\begin{figure*}[t]
  \centering
   \includegraphics[width=\linewidth]{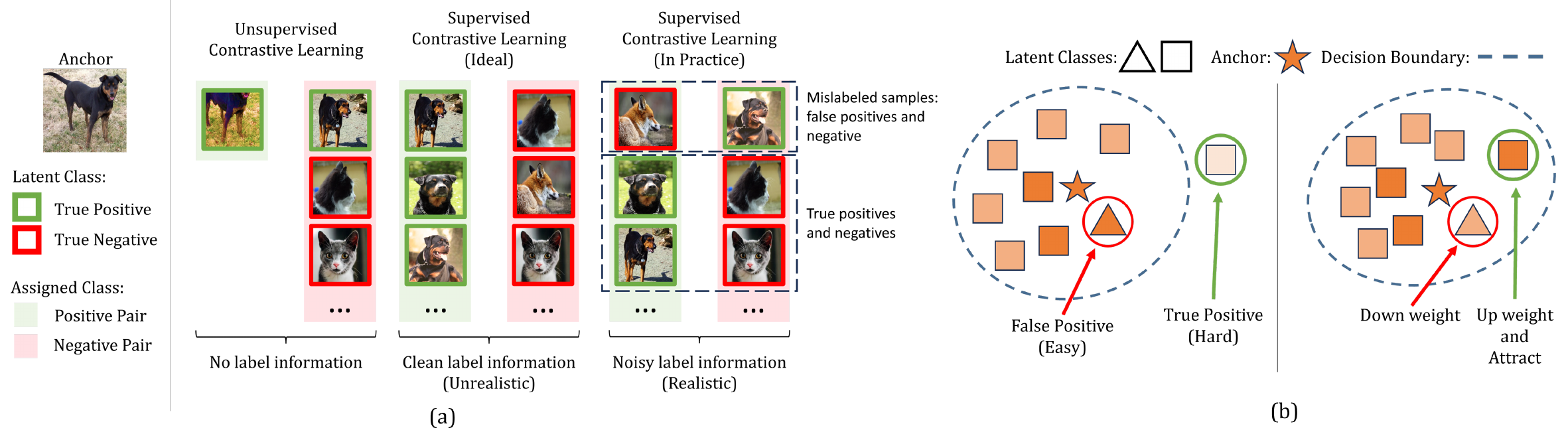}
   \caption{\textbf{(a)} Comparison between different contrastive learning approaches. Supervised learning with noisy labels reflects real-world scenarios. Sample pairs are categorized as true positives or negatives if correctly labeled, and as false positives or negatives if mislabeled.
   \textbf{(b)} Our method, D-SCL, mitigates the bias from mislabeled samples in the context of the positive set. We use shapes to indicate latent class labels and color shade to represent assigned weights during learning. D-SCL decreases the weight of false positives and increases that of true positives to enhance separability of classes 
   and create tighter clusters.}
   \label{fig:CL_compare}
   \vspace{-4mm}
\end{figure*}

Existing approaches that aim to mitigate noisy labels in contrastive learning typically employ techniques such as specialized selection pipelines \cite{DBLP:journals/corr/SongXJS15, DBLP:journals/corr/SchroffKP15}, including bilevel optimization based on the principles of cross-validation \cite{DBLP:conf/eccv/JenniF18}, regularized estimation of annotator confusion matrices \cite{DBLP:conf/cvpr/TannoSSAS19} and integrating open-set samples with dynamic noisy labels \cite{DBLP:conf/nips/WeiTXA21}.
However, these are limited by their sensitivity to hyperparameters and their complex architectures \cite{DBLP:journals/tnn/SongKPSL23,DBLP:conf/cvpr/YaoSZS00T21}.
Other approaches assign a confidence value to each pair and learn largely from the confident pairs \cite{DBLP:conf/icml/MaWHZEXWB18,DBLP:conf/icml/ZhengWG0MC20,DBLP:conf/aaai/ChenYCZH21,li2022selective,huang2023twin}. However, these strategies risk overfitting to potentially incorrect labels or a subset of training data, particularly in datasets with many similar classes \cite{DBLP:journals/tnn/SongKPSL23,DBLP:conf/cvpr/YaoSZS00T21}. They can also introduce significant computational complexity and overhead. For example, Sel-CL \cite{li2022selective} is challenging to apply on large datasets since it uses the $k$-NN algorithm to create pseudo-labels.

In our study, we analyze the effect of noisy labels on Supervised Contrastive Learning (SCL). Our empirical examination, leveraging label data verified by Northcutt et al. \cite{DBLP:conf/nips/NorthcuttAM21} (Sec.~\ref{sec:label_analysis}), reveals a notable similarity overlap in the representation distributions of both correctly labeled and mislabeled samples. This indicates that mislabeled samples, often indistinguishable in representation space, contribute to classification challenges in SCL. Our analysis shows that in approximately 99\% of cases, the performance decline in SCL is attributed to incorrect learning signals from mislabeled samples in the positive set (false positives). 



Based on this analysis, we propose a novel sampling strategy that emphasises true positives, which come from the same latent class but are distantly placed, and true negatives with similar representations (Sec.~\ref{sec:method}). In contrast, existing noise-robust methods \cite{DBLP:conf/icml/ArazoOAOM19,DBLP:conf/sisap/Houle17} typically assign greater weight to confident pairs that are closely positioned, despite the high likelihood of these pairs being false positives. Furthermore, based on the established benefits of utilizing `hard' negative samples \cite{DBLP:journals/corr/SongXJS15, DBLP:journals/corr/SchroffKP15,DBLP:conf/iclr/RobinsonCSJ21}, we hypothesise that \textit{true positives originating from the same latent class, yet positioned distantly, could be important for enhancing the quality of learned representations}. As a result, we strategically reduce the risk of an incorrect decision boundary, as shown in Fig.~\ref{fig:CL_compare}. In summary, our main contributions are:



\begin{itemize}
\item \textbf{Insights into the impact of noisy labels on supervised contrastive learning}: 
We present an in-depth analysis in Sec.~\ref{sec:label_analysis}, elucidating the impact of mislabeled samples on SCL and offering strategies for their effective mitigation.

\item \textbf{A novel technique for efficient, noise-robust contrastive learning}: We propose \textit{D-SCL} in Sec.~\ref{sec:method}, a novel SCL objective that is the first to remove bias due to misannotated labels within a supervised contrastive framework. Unlike previous works \cite{DBLP:conf/icml/ZhengWG0MC20,DBLP:conf/aaai/ChenYCZH21,li2022selective,huang2023twin}, 
D-SCL does not introduce extra computational overhead and is suitable for general image benchmarks with large image models, thereby optimizing performance without sacrificing processing speed or resource utilization.
\item \textbf{State-of-the-art performance}: Compared to traditional pre-training using cross-entropy, D-SCL achieves significant gains in Top-1 accuracy across multiple datasets including iNat2017 \citep{DBLP:conf/cvpr/inat2017}, ImageNet \citep{DBLP:conf/cvpr/imagenet}, and others (Sec.~\ref{sec:exp}). Furthermore, when using existing pre-trained weights, D-SCL demonstrates superior performance on transfer learning, outperforming existing methods such as Sel-CL \cite{li2022selective} and setting a new state-of-the-art for base models (with 88M parameters) on ImageNet-1K. 
\end{itemize}


\section{Related Work}

Label noise is prevalent in many datasets used for supervised learning, especially for large datasets where noise removal is impractical \cite{DBLP:conf/nips/NorthcuttAM21,DBLP:journals/corr/abs-2208-08464}. Mislabeled examples can lead to overfitting in models, with larger models being more susceptible than smaller ones \cite{rethinkgeneralization,DBLP:journals/tnn/SongKPSL23,DBLP:journals/corr/abs-2011-04406,DBLP:conf/iccS/DubelWN23, DBLP:conf/nips/NorthcuttAM21}. Robust learning from noisy labels is thus crucial for improving generalization. Methods include estimating noise transition matrices \cite{holviewtransmat, transmat2, dual-t-transmatrix}, regularization \cite{DBLP:journals/corr/abs-1710-09412, RegularizeNoise, Bilevel}, and sample re-weighting \cite{DBLP:conf/icml/RenZYU18,DBLP:journals/tnn/WangLT18}. The influence of noisy labels in supervised contrastive learning (SCL) is less explored, which this paper addresses by developing noise-mitigating methods.

Contrastive learning, particularly in unsupervised visual representation learning, has evolved significantly \cite{becker1992self, RN89, mocov2, barlowtwins, byol, swav}. However, the absence of definitive label information can result in samples within negative pairs actually belonging to the same latent class, potentially leading to detrimental effects on the representations learned. Supervised contrastive learning (SCL) leverages labeled data to construct positive and negative pairs based on semantic concepts of interest (e.g.~object categories). It ensures that semantically related points are attracted to each other in the embedding space.  Khosla et al.~\cite{RN81} introduce a SCL objective inspired by InfoNCE, which can be considered a supervised extension of previous contrastive objectives---e.g. triplet loss \cite{DBLP:journals/corr/SchroffKP15} and N-pairs loss \cite{npairloss}). Despite its efficacy, the impact of label noise and the importance of hard sample mining in SCL are often overlooked. We build on their work by devising a strategy that not only mitigates the impact of label noise but also enhances the robustness of the contrastive learning objective by explicitly hard sampling.

Several recent works aim to tackle noisy labels in SCL. Ortego et al.~\cite{DBLP:conf/cvpr/OrtegoAAOM21} introduce an adaptation of Mixup augmentation \cite{DBLP:journals/corr/abs-1710-09412}, combined with noisy sample identification, to create a semi-supervised, noise-robust contrastive learning objective. MoPro \cite{MoPro} employs momentum prototypes of classes, utilizing the similarity between a sample’s latent representation and these prototypes to generate pseudo labels, thereby enhancing resilience against mislabeled examples. Li et al.~\cite{SSCL} add a non-linear projection head to compute intra-sample similarities, which helps select confident examples and pairs.
However, these approaches, while innovative, face significant challenges in computational efficiency and application on standard image benchmarks---for example TCL \cite{huang2023twin} is slow since it employs a Gaussian mixture model with entropy-regularized cross-supervision. 
Such approaches also risk overfitting on confident pairs, which can limit their generalizability to new data. To counteract overfitting, they often resort to smaller models (e.g.~ResNet-18), contrasting the common practice of using larger models for higher performance. Furthermore, they predominantly focus on highly noisy datasets, with noise rates up to 80\%, an uncommon scenario in practice. In contrast, our approach, tailored to real-world conditions, specifically addresses human annotation errors. We mitigate the impact of noisy labels by deprioritizing easy positives and reducing the bias from labelling errors without incurring additional computational expense.

\section{Background: Contrastive Learning}

We begin by discussing the fundamentals of contrastive representation learning. Here, the objective is to contrast pairs of data points that are semantically similar (positive pairs) against those that are dissimilar (negative pairs). Mathematically, given a data distribution \( p(x) \) over \( \mathcal{X} \), the goal is to learn an embedding \( f: \mathcal{X} \rightarrow \mathbb{R}^{d} \) such that similar pairs \( (x, x^{+}) \) are close in the feature space, while dissimilar pairs \( (x, x^{-}) \) are more distant. In unsupervised learning, for each training datum \( x \), the selection of \( x^{+} \) and \( x^{-} \) is dependent on \( x \). Typically, one positive example \( x^{+} \) is generated through data augmentations and \( N \) negative examples \( x^{-} \).
The contrastive loss, named InfoNCE or the $N$-pair loss \cite{gutmann2010noise,sohn2016improved,oord2018representation}, is then defined as
\begin{equation}
\resizebox{0.43\textwidth}{!}{$
\mathcal{L}_{\mathrm{NCE}} = \mathbb{E}_{\substack{x\\ x^{+}\\ \left\{x_i^{-}\right\}_{i=1}^N}}\left[-\log \frac{e^{f(x)^T f(x^{+})}}{e^{f(x)^T f(x^{+})}+\sum_{i=1}^N e^{f(x)^T f(x_i^{-})}}\right]
$}
\label{eq:infoloss}
\end{equation}
Here, the expectation computes the average loss across all possible choices of positive and negative samples within the dataset. In practice, during a training iteration, one typically samples a mini-batch; then, for each data point in it (referred to as an `anchor'), a positive example is selected---usually an augmented version of the anchor or another instance of the same class---while the rest of the batch is treated as negative examples. This is under the assumption that within the batch, instances of different classes (i.e., all other samples except the positive pair) serve as negatives.




Khosla \textit{et al.}~\cite{RN81} extended this concept to supervised contrastive learning, experimenting with two losses:
%
%
%
\begin{equation}
\resizebox{0.48\textwidth}{!}{$
\mathcal{L}_{\text {in }}^{ \text {sup} } =
\mathbb{E}_{\substack{x\\ \left\{x_k^{+}\right\}_{k=1}^K \\\left\{x_i^{-}\right\}_{i=1}^N}}
 -\log \left\{\frac{1}{|K|} \sum_{k=1}^K \frac{\exp \left(f(x)^T f\left(x_k^{+}\right)\right)}{\sum_{k=1}^{K} e^{f(x)^T f( x_{k}^{+})}+\sum_{i=1}^N e^{f(x)^T f(x_i^{-})}}\right\}
$}
\label{eq:supercon_in} 
\vspace{-4mm}
\end{equation}
%
%

\begin{equation}
\resizebox{0.48\textwidth}{!}{$
\mathcal{L}_{\text {out }}^{\text {sup }} = 
\mathbb{E}_{\substack{x\\ \left\{x_k^{+}\right\}_{k=1}^K \\\left\{x_i^{-}\right\}_{i=1}^N}}
\frac{-1}{|K|} \sum_{k=1}^K \left[ 
    \log \left\{  
        \frac{
            \exp \left(f(x)^T f\left(x_k^{+}\right)\right)
        }{
            \sum_{k=1}^{K} e^{f(x)^T f( x_{k}^{+})}+\sum_{i=1}^N e^{f(x)^T f(x_i^{-})}
        }
    \right\}
    \right]
$}
\label{eq:supercon_out}
\end{equation}
Here $k$ indexes a set of $K$ positive samples, i.e.~images $x_k^+$ of the same class as $x$.
We focus on improving these objectives, making them more robust to labeling errors in Sec.~\ref{sec:method}.

\section{Analysing Label Errors and their Impacts}
\label{sec:label_analysis}

Label errors are common in benchmark datasets, and can significantly influence the results of model training and evaluation \cite{DBLP:conf/iclr/AdebayoHYC23,DBLP:conf/nips/NorthcuttAM21,DBLP:journals/coling/KlieWG23}.
Despite measures such as having multiple raters annotate each image, 
Northcutt \textit{et al.}~\cite{DBLP:conf/nips/NorthcuttAM21} found that the large-scale ImageNet and CIFAR-100 image datasets have validation-set error rates of 5.83\% and 5.85\%, respectively.
As training sets are less scrutinized, their error rates may be higher, affecting both training and evaluation.

In the following subsections, we investigate the impact of such human labeling errors on supervised contrastive learning (SCL). 
We categorize all scenarios in which mislabeled samples can occur in SCL, identify the most important cases, and consider their adverse effects.
We also study similarities across various sample pairs, finding that
mislabeled samples exhibit high similarity to true positive pairs.

\paragraph{Definitions.}
We define the \textit{latent label} of an image as being its true category (e.g.~the latent label of an image of a cat would be `cat').
The term \textit{assigned label} refers to the class that a human annotator has assigned to an image (hopefully---but not always---matching the latent label).
Given a pair of images,
we define a \textit{false positive} as being when an annotator has erroneously grouped those images under the same assigned label, even though their latent labels are different.
A \textit{false negative} is when two images sharing the same latent label mistakenly have different assigned labels.
We define \textit{true positive} and \text{true negative} pairs analogously.
Lastly, we define \textit{easy positives} as pairs of images that share the same assigned labels and also have highly similar embeddings.


\subsection{Impact of Labelling Error on Constructing True Pairs}
\label{sec:analysis_on_truepair}

In this section, we analyse the interaction of labelling error and supervised contrastive learning (SCL) by looking at the probability of false positives and false negatives during training.
%
Specifically, we hypothesise that we can focus our efforts on mitigating false positives, due to the infrequency of false negatives. To support this hypothesis, we present the following analysis.

\begingroup
\setlength{\columnsep}{4pt}%
\setlength{\intextsep}{4pt}%
\begin{wrapfigure}{r}{0.47\linewidth}
\vspace{-12pt}
    \includegraphics[width=\linewidth]{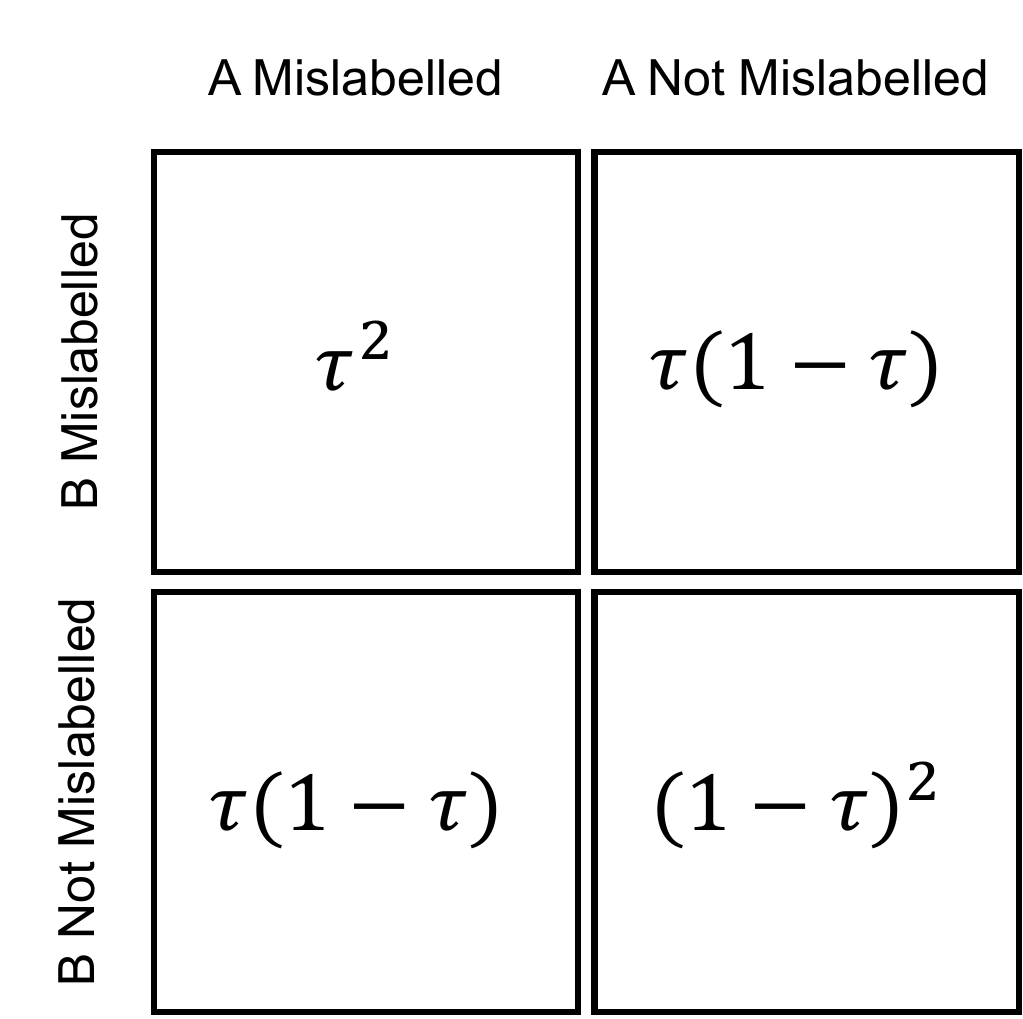}
\end{wrapfigure}
Let `A' represent an anchor image and `B' represent another paired image. The inset figure shows the probabilities of different labelling outcomes under symmetric label noise with an error rate of $\tau$ (i.e.~the probability that any sample is mislabelled is $\tau$). If `A' and `B' are assigned the same label, they are a false positive if: `A' is correctly labeled while 'B' is not; \textit{or} 'A' is mislabeled while 'B' is correctly labeled; \textit{or} `A' and 'B' are mislabeled and do not 
belong to the same latent class.

\endgroup

Conversely, if `A' and `B' are assigned different labels, they are a false negative if both `A' and 'B' are mislabeled but actually belong to the same latent class; \textit{or} `A' is correctly labeled and `B' is not, yet `B' belongs to the same latent class as `A'; \textit{or} `A' is mislabeled and `B' is not, with `A' being of the same latent class as `B'.

Therefore, we can estimate the probabilities of false positive and negatives in a minibatch using the error rate estimates from \cite{DBLP:conf/nips/NorthcuttAM21}. 
For instance, in CIFAR-100 (5.85\% error rate and 100 classes), about 11.4\% of positive pairs and roughly 0.11\% of negative pairs are likely to be falsely labeled.
In ImageNet (5.83\% error rate and 1000 classes), we anticipate approximately 11.32\% of positive pairs and about 0.09\% of negative pairs to be incorrect.
Based on this observation, we argue that when tackling labeling errors in contrastive learning, we can largely ignore false negatives due to their very low likelihood.
We provide estimates of these rates for other datasets in the supplementary material.

\subsection{The Similarity Between Different Pairs}
\label{sec:similarity_of_mis}

\begin{figure}[t]
    \centering
    \includegraphics[width=\linewidth]{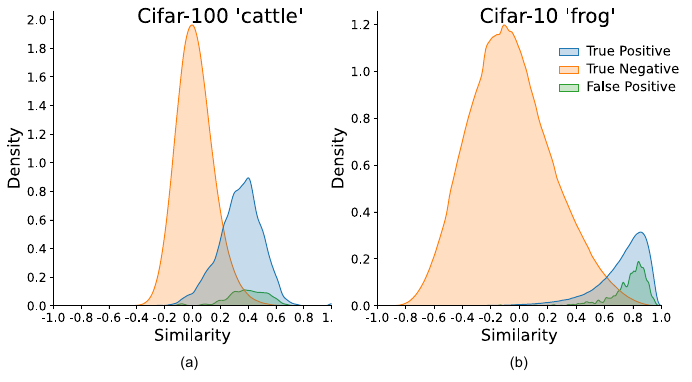}
    \caption{Distributions of similarities for true positive pairs, true negative pairs and false positives (mislabeled samples paired with samples from the assigned label class), for two classes of anchor.}
    \label{fig:similarity map}
    \vspace{-4mm}
\end{figure}

Based on the conclusion from Sec.~\ref{sec:analysis_on_truepair} that we should focus on false positives, we now conduct an in-depth empirical exploration of these. We hypothesize that false positive samples are grouped with the positive set in relation to the anchor due to their similar visual characteristics, making accurate labelling difficult for both humans and machines.
This aligns with the concept of `easy positives'. We evaluate this hypothesis using standard datasets to gain further understanding. This insight leads us to the strategy of reducing the weighting of easy positives as an efficient method to diminish the influence of false positives. 

We begin by analysing the quality of embeddings of different pairs. We use the CIFAR-10, CIFAR-100 and ImageNet-1k datasets, and ViT-base models pretrained on each separately. We use the consensus among annotators from \cite{DBLP:conf/nips/NorthcuttAM21} to determine the latent class of samples.
%
Fig.~\ref{fig:similarity map} shows the relationship between similarities of different types of pairs for two image classes. These plots depict the distribution of cosine similarities by comparing embeddings across three distinct categories: true positive pairs, true negative pairs, and false positive pairs (mislabeled samples paired with samples from the assigned label class). These plots support our hypothesis that false positive samples are `easy positive' samples. With close inspection of Fig.~\ref{fig:similarity map}, we can observe a noticeable overlap in the similarity distributions of true positives and false positives, showing a high similarity in their embeddings. This overlap is notably greater than that with true negatives, underlining a strong association between true and false positives. Further results on other datasets are given in the supplementary material, as well as a quantitative evaluation.
Overall, both qualitative and quantitative results show that false positives and true positives have high similarity. This not only validates our hypothesis but also provides support for our method, which incorporates less weighting on easy positives and reduces bias caused by labeling errors.

\section{Debiased Contrastive Learning}
\label{sec:method}

In this section, we describe our approach to reduce bias caused by noisy labels in positive pairs and how this fits into an overall contrastive learning objective. 
In Sec.~\ref{sec:label_analysis}, we noted that the most pronounced impact of the mislabeled samples arises when they are incorporated into the positive set and exhibit high similarity to the anchor (i.e.~are easy positives). 
%
Our method therefore adheres to two key principles (Fig.~\ref{fig:CL_compare}b): (\textbf{P1}) it should ensure that the \textit{latent} class of positive samples matches that of the anchor~\citep{RN89,RN91,DBLP:journals/corr/abs-2007-00224,DBLP:journals/corr/abs-2010-01028}; and (\textbf{P2}) it should deprioritize easy positives, i.e.~those currently embedded near the anchor.
By reducing the weighting of easy positives, we minimize the effect of wrong learning signals from false positive pairs. The model is also forced to recognize and encode deeper similarities that are not immediately apparent, improving its discriminative ability.

\subsection{Label Noise in Supervised Contrastive Learning}


Khosla~\textit{et al.}~\cite{RN81} argued that $\mathcal{L}_{\text{out}}^{\text{sup}}$ is superior to $\mathcal{L}_{\text{in}}^{\text{sup}}$, attributing this to the normalization factor $\frac{1}{|P(i)|}$ in $\mathcal{L}_{\text{out}}^{\text{sup}}$ that mitigates bias within batches. Although $\mathcal{L}_{\text{in}}^{\text{sup}}$ incorporates the same factor, its placement inside the logarithm reduces its impact to a mere additive constant, not influencing the gradient and leaving the model more prone to bias.
We instead introduce a modified $\mathcal{L}_{\text{in}}^{\text{sup}}$ that directly reduces bias due to mislabeling,
and outperforms $\mathcal{L}_{\text{out}}^{\text{sup}}$.

We begin with a modified formulation of $\mathcal{L}_{\text{in}}^{\text{sup}}$ (Eq.~\ref{eq:supercon_in}), that is equivalent up to a constant scale and shift, but will prove easier to adapt:
%
\begin{equation}
\resizebox{0.43\textwidth}{!}{$
\mathbb{E}_{\substack{x\\ \left\{x_k^{+}\right\}_{k=1}^K \\\left\{x_i^{-}\right\}_{i=1}^N}}  \left[  \log \frac{-1}{\left |K  \right |}  \frac{\sum_{k=1}^{K}     e^{f(x)^T f(x_{k}^{+})}}{\sum_{k=1}^{K} e^{f(x)^T f( x_{k}^{+})}+\sum_{i=1}^N e^{f(x)^T f(x_i^{-})}}\right]
$}
\label{eq:our_supercon} 
\end{equation}
Here all \( K \) samples from the same class within a mini-batch are treated as positive samples for the anchor $x$.

We now introduce our main technical contribution, which is a debiased version of (\ref{eq:our_supercon}) that is more robust to label noise.
As in Sec.~\ref{sec:label_analysis}, we assume there is set of latent classes \( \mathcal{C} \), that encapsulate the semantic content, and hopefully match the assigned labels.
%
Following \cite{DBLP:journals/corr/abs-1902-09229,DBLP:journals/corr/abs-2007-00224,DBLP:conf/iclr/RobinsonCSJ21}, pairs of images \( (x, x^{+}) \) are supposed to belong to the same latent class $c$, where $x \in \mathcal{X}$ is drawn from a data distribution $p(x)$.
Let $\tau$ denote the probability that any sample is mislabeled; we assume this is constant for all $x$. 
Since $\tau$ is unknown in practice, it must be treated as hyperparameter, or estimated based on previous studies.
We also introduce an (unknown) function \( z: \mathcal{X} \rightarrow \mathcal{C} \) that maps $x$ to its latent class label.
Then, \( p_{x}^{+} := p(x' \,|\, z(x') = z(x)) \) is the probability of observing \( x' \) as a positive example for \( x \), whereas \( p_{x}^{-} = p(x' \,|\, z(x') \neq  z(x)) \) is the probability of a negative.
For each image \( x \), the objective (Eq.~\ref{eq:our_supercon}) aims to learn a representation \( f(x) \) by using positive examples \( \{x^{+}\}_{k=1}^{K} \)  with the same latent class label as \( x \) and negative examples \( \{x_{i}^{-}\}_{i=1}^{N} \) that belong to different latent classes.
Since $p$ is the true prior distribution, the ideal loss function to be minimized (c.f.~\cite{DBLP:journals/corr/abs-2007-00224}) is:
%
%
\begin{equation}
\resizebox{0.45\textwidth}{!}{$
\mathcal{L}_{T}  = \mathbb{E}_{\substack{x\sim p \\ x_{k}^{+} \sim  p_{x}^{+} \\ x_{i}^{-} \sim  p_{x}^{-}}}   \left[   \frac{-1}{\left |K  \right |}       \log \frac{\frac{Q}{K}  \sum_{k=1}^{K}     e^{f(x)^T f(x_{k}^{+})}}{\frac{Q}{K} \sum_{k=1}^{K} e^{f(x)^T f( x_{k}^{+})}+ \frac{W}{N}\sum_{i=1}^N e^{f(x)^T f(x_i^{-})}}\right]
$}
\label{eq:supercon_weighted} 
\end{equation}
%
We term this loss function the \textit{true label loss}. Here \( Q \)  and \( W \) are weighting parameters that we introduce to help with analysing the bias; when they equal the numbers of positive and negative examples respectively, $\mathcal{L}_T$ reduces to the conventional supervised contrastive loss (\ref{eq:our_supercon}). Note that supervised contrastive learning typically assumes \( p_{x}^{+} \) and \( p_{x}^{-} \) can be determined from human annotations; however since we consider latent classes instead of assigned classes, we do \textit{not} have access to the true distribution.
However, we now show how to approximate this true distribution more closely and improve the overall performance.

\subsection{Debiasing the Contrastive Loss}


For a given anchor $x$ and its embedding $f(x)$, we now aim to build a distribution $q$ on $\mathcal{X}$ that fulfils the principles \textbf{P1} and \textbf{P2}.
We will draw a batch of positive samples \( \{ x_{k}^{+} \}_{k=1}^{K} \) from \( q \).
Ideally we would draw samples from
%
\begin{align}
    q^{+} (x^{+}) := \,& q(x^{+}\,|\,z(x) = z(x^{+})) \\
    \propto \,& \frac{1}{e^{\beta f(x)^{T}f(x^{+})}} \cdot p(x^{+})
    \label{eq:q}
\end{align}
where $\beta \geq 0$. It is important to note that \( q^{+}(x^+) \) depends on \( x \), although this dependency is not explicitly shown in the notation.
The distribution is composed of two factors:
\begin{itemize}
    \item The event \( \{z(x) = z(x^{+}) \} \) indicates that pairs, \( (x, x^{+}) \), should originate from the same latent class (\textbf{P1}). $p^+_x(x^{+})$ is the true (unknown) positive distribution for anchor $x$.
    \item 
    The exponential term is an unnormalized von Mises-Fisher distribution with a mean direction of \( f(x) \) and a concentration parameter \( \beta \). 
    This term increases the probability of sampling hard positives, (\textbf{P2}).
    The concentration parameter \( \beta \) modulates the weighting scheme of \( q^{+} \), specifically augmenting the weights of instances \( x^{+} \) that exhibit a lower inner product (i.e.~greater dissimilarity) with the anchor point \( x \). 
\end{itemize}

The distribution \( q^{+} \) fulfils our desired principles of selecting true positives and deprioritizing easy positives.
However, we do not have the access to the latent classes, and so cannot directly sample from it.
We therefore rewrite it from the perspective of Positive-Unlabeled (PU) learning \cite{elkan2008learning,du2014analysis,DBLP:conf/iclr/RobinsonCSJ21,RN126}, which will allow us to implement an efficient sampling mechanism. 
We first define \( q^{-}(x^{+}) \propto \frac{1}{e^{\beta f(x)^{T}f(x^{+})}} \cdot p^{-}(x^{+}) \).
Then, by conditioning on the event \( \{ z(x) = z(x^{+}) \} \), we can write 
%
%
%
\begin{align}
    q (x^{+}) &= \tau^{+}q^{+}(x^{+}) +\tau^{-}q^{-}(x^{+})
    \label{eq:pos_split}
\\ \Rightarrow \;
 q^{+}(x^{+})  &=  \left( q (x^{+}) - \tau^{-}q^{-}(x^{+}) \right) / \tau^{+}
 \label{eq:pos_rerrange}
\end{align}
where $\tau^{+}$ is the probability that a sample from the data distribution $p(x)$ will have the same latent class as $x$.  

\looseness -1 We have now derived an expression (\ref{eq:pos_rerrange}) for the positive sampling distribution \( q^{+} \), expressed in terms of two tractable distributions that depend only on the assigned (not latent) labels. 
Importantly, sampling from \( p \) does \textit{not} require knowledge of true latent classes. To sample from \( q\) and \( q^{-}\) by importance sampling, we first hold the weighting parameter \( Q \) fixed and consider a sufficiently large value for  \( K \), the number of positive samples for the anchor $x$. Then, the supervised contrastive learning objective (Eq. \ref{eq:supercon_weighted}) becomes: 
%
%
\begin{equation}
\resizebox{0.43\textwidth}{!}{$
L_{T}  = \mathbb{E}_{\substack{x\sim p  \\ x^{-} \sim  p_{x}^{-}}}  \left[ \log \frac{-1}{\left |K  \right |}  \frac{Q  \mathbb{E}_{x^{+} \sim q}\left[e^{f(x)^T f(x^{+})}\right] }{Q  \mathbb{E}_{x^{+} \sim q}\left[e^{f(x)^T f\left(x^{+}\right)}\right] + \frac{W}{N}\sum_{i=1}^N e^{f(x)^T f(x_i^{-})}}\right]
    $}
\label{eq:positive_sampling} 
\end{equation}
%


By substituting Eq. \ref{eq:pos_rerrange} into Eq. \ref{eq:positive_sampling}, we obtain an objective that removes bias from labeling errors and also down-weights easy positives: 
%
%
%

%
\begin{equation}
\resizebox{0.48\textwidth}{!}{$
\mathbb{E}_{\substack{x\sim p\\ x^{+} \sim  q \\ x^{-} \sim  q}}   \left[ \log \frac{-1}{\left |K  \right |}  \frac{\frac{Q}{\tau^{+}}\left(\mathbb{E}_{x^{+} \sim q}\left[e^{f(x)^T f(x^{+})}\right]-\tau^{-} \mathbb{E}_{v \sim q^{-}}\left[e^{f(x)^T f(v)}\right]\right)}    
{\frac{Q}{\tau^{+}}\left(\mathbb{E}_{x^{+} \sim q}\left[e^{f(x)^T f(x^{+})}\right]-\tau^{-} \mathbb{E}_{v \sim q^{-}}\left[e^{f(x)^T f(v)}\right]\right)   +  \frac{W}{N}\sum_{i=1}^N e^{f(x)^T f(x_i^{-})}}\right]
   $}
\label{eq:hard_pos}
\end{equation}
This suggests we only need to approximate the expectations $\mathbb{E}_{x^{+} \sim q}\left[e^{f(x)^T f\left(x^{+}\right)}\right]$ and $\mathbb{E}_{v \sim q^{-}}\left[e^{f(x)^T f(v)}\right]$ over \( q \) and \( q^{-} \), which can be achieved by classical Monte Carlo importance sampling, using samples from \( p \) and \( p^{-} \):
%
\begin{equation}
\resizebox{0.48\textwidth}{!}{$
\mathbb{E}_{x^{+} \sim q}\left[e^{f(x)^T f\left(x^{+}\right)}\right]=\mathbb{E}_{x^{+} \sim p}\left[e^{f(x)^T f(x^{+})} q / p\right]=\mathbb{E}_{x^{+} \sim p}\left[e^{(\beta+1) f(x)^T f(x^{+})} / Z(x)\right]
$}
\label{eq:monte_pos}
\end{equation}
\begin{equation}
\resizebox{0.48\textwidth}{!}{$
\mathbb{E}_{v \sim q^{-}}\left[e^{f(x)^T f(v)}\right]=\mathbb{E}_{v \sim p^{-}}\left[e^{f(x)^T f(v)} q^{-} / p^{-}\right]=\mathbb{E}_{v \sim p^{-}}\left[e^{(\beta+1) f(x)^T f(v)} / Z^{-}(x) \right]
$}
\label{eq:monte_pos_v}
\end{equation}
where, \( Z(x) \) and \( Z^{-}(x) \) are the partition functions for \( q \) and \( q^{-} \) respectively. Hence, these expectations over \( p \) and \( p^{-} \) admit empirical estimates
%
\begin{align}
    \widehat{Z}(x) &=\frac{1}{M} \sum_{i=1}^M e^{\beta f(x)^{\top} f(x_i^{+})}
    \\
    \widehat{Z}^{-}(x) &=\frac{1}{N} \sum_{i=1}^N e^{\beta f(x)^{\top} f(x_i^{-})}
\vspace{-3mm}
\end{align}


\paragraph{Debiasing negatives.}
%
%
Despite the minimal impact of mislabeled samples in negative sets (see Sec.~\ref{sec:label_analysis}), we extend our debiased method to these samples to further reduce their adverse effects. Mirroring our positive sample strategy, the debiasing process for negatives involves constructing a distribution that not only aligns with true negatives but also places greater emphasis on hard negatives.
We then use Monte Carlo importance sampling techniques to better estimate the true distribution of latent classes. Full details are given in the supplementary material. 

\begin{figure*}
\small
\centering
\begin{equation}
\mathbb{E}_{\substack{x\sim p\\ x^{+} \sim  q \\ x^{-} \sim  q}} \left[ \log \frac{-1}{\left |K  \right |}  \frac{\frac{Q}{\tau^{+}}\left(\mathbb{E}_{x^{+} \sim q}\left[e^{f(x)^T f(x^{+})}\right] - \tau^{-} \mathbb{E}_{v \sim q^{-}}\left[e^{f(x)^T f(v)}\right]\right)}{\frac{Q}{\tau^{+}}\left(\mathbb{E}_{x^{+} \sim q}\left[e^{f(x)^T f(x^{+})}\right] - \tau^{-} \mathbb{E}_{v \sim q^{-}}\left[e^{f(x)^T f(v)}\right]\right) + \frac{W}{\tau^{-}}\left(\mathbb{E}_{x^{-} \sim q}\left[e^{f(x)^T f(x^{+})}\right] - \tau^{+} \mathbb{E}_{b \sim q^{+} }\left[e^{f(x)^T f(b)}\right]\right)}\right]
\label{eq:easy_contrastivelearning}
\end{equation}
\vspace{-7pt}
\end{figure*}

\paragraph{Overall learning objective.}
Using Eq.~\ref{eq:hard_pos} and combining debiasing for negatives, we get Eq.~\ref{eq:easy_contrastivelearning}.
This objective has the following desirable properties:
\begin{itemize}

\looseness -1 \item \textbf{Mitigates the adverse impact of mislabelled samples}: 
Given the analysis in Sec.~\ref{sec:label_analysis}, it is critical to reduce the adverse impact of false positives (mislabelled samples). The embeddings of mislabelled samples are often very close to the anchor, making them more likely to be easy positives. By effectively giving less weight to easy positives (mislabelled samples), thereby reducing their impact on providing misleading learning signals. Beyond sophisticated weighting, we also reduce the bias due to noisy labels, assuming access only to noisy labels. Specifically, we develop a correction for the mislabelled sample bias, leading to a new, modified loss termed debiased supervised contrastive loss (Eq.~\ref{eq:easy_contrastivelearning}). Our approach indirectly approximates the true latent distribution, which prevents contrastive learning from being misled by incorrectly labeled samples and also reinforces the core philosophy of contrastive learning. 
 
\looseness -1 \item \textbf{Discriminating fine detail with hard samples}: Our methodology adjusts the weighting of all samples based on their “hardness”. This nuanced approach ensures that the model differentiates not only between distinctly different samples but also hones its skills on more challenging, closely related negative samples. Such an approach paves the way for a robust model that discerns in real-world scenarios where class differences might be minimal.
\end{itemize}

\section{Experiments}
\label{sec:exp}

We evaluate our proposed method, D-SCL, on image classification in three settings: pre-training from scratch, transfer learning using pre-trained weights, and pre-training with highly noisy datasets. We also conduct several ablation experiments. For all experiments, we use the official train/test splits and report the mean Top-1 test accuracy across three distinct initializations.

We employ representative models from two categories of architectures -- BEiT-3/ViT base \citep{beit3,rn83}, and ResNet-50 \cite{RN36}.
While new state-of-the-art models are continuously emerging (e.g.~DINOv2 \cite{DBLP:journals/corr/abs-2304-07193}), our focus is not on the specific choice of architecture. Instead, we aim to show that D-SCL is model-agnostic, and enhances performance on two very different architectures.
Further implementation details and the complete code for all experiments can be found in the supplementary material, which will be made publicly available upon acceptance.

\subsection{Training from Scratch}
\label{subsec:results-pretrain}

\begin{table}[t]
\centering
\resizebox{\linewidth}{!}{
\scriptsize
\begin{tabular}{@{}l|cccc@{}}
\toprule
\multicolumn{1}{@{}l|}{\textbf{Model}} &\multicolumn{1}{@{}c}{\textbf{Loss}} &\multicolumn{1}{@{}c}{\textbf{CIFAR-10}}& \multicolumn{1}{@{}c}{\textbf{CIFAR-100}}& \multicolumn{1}{@{}c}{\textbf{ImageNet}} \\
\midrule
\multirow{5}{*}{\textbf{BEiT-3}}&CE\cite{DBLP:journals/bstj/Shannon48}&71.70&59.67&77.91\\ 

                                &SupCon\cite{RN81}&88.96&60.77&82.57\\ 

                                &Sel-CL\cite{li2022selective}&86.33&59.51&81.87\\ 

                                &TCL\cite{huang2023twin}&85.16&59.22&81.74\\ 

                                &\multicolumn{1}{@{}c}{D-SCL (ours)}&\textbf{90.16}&\textbf{64.47}&\textbf{84.21}\\ 
\midrule
\multirow{5}{*}{\textbf{ResNet-50}}&CE\cite{DBLP:journals/bstj/Shannon48}&95.00&75.30&78.20\\ 

                                &SupCon\cite{RN81}&96.00&76.50&78.70\\ 

                                &Sel-CL\cite{li2022selective}&93.10&74.29&77.85\\ 

                                &TCL\cite{huang2023twin}&92.80&74.14&77.17\\ 

                                &D-SCL (ours)&\textbf{96.39}&\textbf{77.82}&\textbf{79.15}\\ 
\bottomrule
\end{tabular}}
\caption{Model accuracy measured using the acc@1 metric when trained with different loss functions on 3 popular image classification benchmarks. All models here are trained from scratch using only the indicated dataset, without pretraining. }
\label{table:scratch}
\vspace{-3mm}
\end{table}


We evaluate our proposed D-SCL objective in the pre-training setting, i.e.~training randomly initialized models from scratch without the use of additional data. Following \cite{RN81}, to use the trained models for classification, we train a linear layer on top of the frozen trained models using a cross-entropy loss. 
We use three benchmarks: CIFAR-10, CIFAR-100 \citep{krizhevsky2009learning}, and ImageNet-1k \citep{DBLP:conf/cvpr/imagenet}.
Tab.~\ref{table:scratch} shows the performance of BEiT-3 and ResNet-50, with different loss functions on three popular image classification datasets. It is noteworthy that, due to the absence of pre-trained weights, BEiT-3 is exactly identical to the ViT model \cite{beit3}. We compare against training with the standard cross-entropy loss as well as the state-of-the-art supervised contrastive learning loss (SupCon) \cite{RN81}. Additionally, we compare against two noise-robust contrastive learning strategies (Sel-CL~\cite{li2022selective} and TCL~\cite{huang2023twin}). We see that D-SCL consistently improves classification accuracy over other training objectives. On Imagenet-1k, D-SCL leads to a 6.3\% and 1.6\% improvement in accuracy for BEiT-3, relative to cross-entropy and SuperCon training, respectively. 

We find that D-SCL outperforms the existing noise-robust contrastive methods Sel-CL and TCL, e.g.~ D-SCL performs 2.4\% and 2.3\% better than Sel-CL for BEiT-3 and ResNet50 on ImageNet, respectively. 
We speculate that due to discarding many pairs, Sel-CL and TCL overfit a subset of training samples, limiting their performance when the rate of label noise is relatively low (e.g. 5.85\% for CIFAR-100 \cite{DBLP:conf/nips/NorthcuttAM21}), and allowing them to be exceeded by SupCon. In contrast, D-SCL outperforms even SupCon, suggesting it is more applicable for real-world image benchmarks with low-to-moderate noise rates.

It is well known that transformer-based models underperform when training data is limited \cite{rn83,rn130,rn131}. This is highlighted by the low performance on CIFAR-10 and CIFAR-100 with cross-entropy training. We show that D-SCL, and to a lesser extent SuperCon, mitigate this---relative to cross-entropy, D-SCL gives a 19\% improvement on CIFAR-10. While BEiT-3 still fails to reach the performance of ResNet-50, the supervised contrastive approaches significantly close the gap, improving the applicability of transformer-based models in limited data scenarios.

In the supplementary material, we also include an ablation study that measures the benefit of different aspects of our method. This shows that mislabelled samples impact the performance of supervised contrastive learning due to their occurrence as soft positives, and that debiasing both positives and negatives helps to improve performance.


\begin{table}[t]
\centering
\resizebox{\linewidth}{!}{
\begin{tabular}{@{}l|l|cccccc@{}}
\toprule
\multicolumn{1}{@{}l}{\textbf{Loss}}& \multicolumn{1}{@{}l}{\textbf{Test set}}& \multicolumn{1}{l}{\textbf{CIFAR-10}} & \multicolumn{1}{l}{\textbf{CIFAR-100}} & \multicolumn{1}{l@{}}{\textbf{ImageNet}} \\
\midrule
\multirow{2}{*}{\textbf{CE}\cite{DBLP:journals/bstj/Shannon48}}     & Original &  \multicolumn{1}{l}{71.70}& \multicolumn{1}{l}{59.67} & \multicolumn{1}{l}{77.91}\\
                       &                             Corrected  & 71.79 \texttt{\small\color{red}(+0.09)}& 59.82 \texttt{\small\color{red}(+0.15)} & 78.24 \texttt{\small\color{red}(+0.33)} \\ \midrule
\multirow{2}{*}{\textbf{SupCon}\cite{RN81}}  & Original  &\multicolumn{1}{l}{ 88.96} &\multicolumn{1}{l}{60.77} &\multicolumn{1}{l}{82.57}   \\
                       &                              Corrected  & 89.11 \texttt{\small\color{red}(+0.15)} & 61.49 \texttt{\small\color{red}(+0.72)} & 83.74 \texttt{\small\color{red}(+1.17)} \\ \midrule
\multirow{2}{*}{\textbf{Sel-CL}\cite{li2022selective}}  & Original &\multicolumn{1}{l}{86.33} &\multicolumn{1}{l}{59.51} &\multicolumn{1}{l}{81.87}  \\
                       &                              Corrected  & 86.21 \texttt{\small\color{blue}(-0.12)} & 58.62 \texttt{\small\color{blue}(-0.89)} & 81.35 \texttt{\small\color{blue}(-0.52)} \\ \midrule
\multirow{2}{*}{\textbf{TCL}\cite{huang2023twin}}  & Original  &\multicolumn{1}{l}{85.16}  &\multicolumn{1}{l}{59.22}  &\multicolumn{1}{l}{81.74}  \\
                       &                             Corrected  & 84.97 \texttt{\small\color{blue}(-0.19)} & 58.14 \texttt{\small\color{blue}(-1.08)} & 81.28 \texttt{\small\color{blue}(-0.46)} \\ \midrule
\multirow{2}{*}{\textbf{D-SCL (ours)}}   & Original  &\multicolumn{1}{l}{90.16}  &\multicolumn{1}{l}{64.47} & \multicolumn{1}{l}{84.21} \\
                       &                              Corrected  &\textbf{90.41} \texttt{\small\color{red}(+0.25)} & \textbf{65.34} \texttt{\small\color{red}(+0.87)} & \textbf{86.02} \texttt{\small\color{red}(+1.81)}  \\ \bottomrule
\end{tabular}}
\caption{Accuracy of the BEiT-3 model using the metric acc@1 on different datasets and with various loss functions, when evaluation on both original and corrected test-set labels.}
\vspace{-1mm}
\label{table:model_accuracy}
\end{table}

\begin{table*}[t]
\centering
\resizebox{17.5cm}{!}{
\begin{tabular}{@{}ll|cccccccc@{}}
\toprule
\textbf{Model} & \textbf{FT method} & \textbf{CIFAR-100} & \textbf{CUB-200} & \textbf{Caltech-256} & \textbf{Oxford-Flowers} & \textbf{Oxford-Pets} & \textbf{iNat2017} & \textbf{Places365} & \textbf{ImageNet-1k} \\ \midrule
ViT            & CE\cite{DBLP:journals/bstj/Shannon48}                 & 87.13              & 76.93            & 90.92               & 90.86                   & 93.81        & 65.26             & 54.06              & 77.91                \\
BEiT-3         & CE\cite{DBLP:journals/bstj/Shannon48}                 & 92.96              & 98.00               & 98.53               & 94.94                   & 94.49        & 72.31             & 59.81              & 85.40                 \\
BEiT-3         & SupCon\cite{RN81}           & 93.15              & 98.23            & 98.66               & 95.10                    & 94.52        & 72.85             & 60.31              & 85.47                \\
BEiT-3         & Sel-CL\cite{li2022selective}           & 91.48              & 94.52            & 97.19               & 93.71                    & 94.51        & 72.43             & 58.36              & 85.21                \\
BEiT-3         & TCL\cite{huang2023twin}           & 90.92              & 93.89            & 97.26               & 93.89                    & 94.68        & 72.47             & 59.22              & 85.18                \\
BEiT-3         & D-SCL (ours)           & \textbf{93.81}              & \textbf{98.95}            & \textbf{99.41}               & \textbf{95.89}                   & \textbf{96.41}        & \textbf{76.25}             & \textbf{62.53}              & \textbf{86.51}                \\ \bottomrule
\end{tabular}}
\caption{Classification accuracy after fine-tuning a pretrained BEiT-3 with different loss functions, on several benchmarks.}
\label{table:transfer}
\vspace{-3mm}
\end{table*}

\paragraph{Performance on corrected test sets.}
%
We next evaluate the same trained models, but using the corrected test-set labels from Northcutt \textit{et al.}~\cite{DBLP:conf/nips/NorthcuttAM21} (Tab.~\ref{table:model_accuracy}).
Importantly, we can observe a relatively larger increase in performance on the corrected test sets with D-SCL---e.g. an improvement of 1.81\% on ImageNet-1k.
This contrasts with SuperCon~\cite{RN81} and cross-entropy, which show a lesser improvement of 1.17\% and 0.33\%, respectively. This supports the claim that D-SCL is less prone to overfitting to label noise than over SuperCon and cross-entropy. 
We find that Sel-CL\cite{li2022selective} and TCL\cite{huang2023twin} generally lead to worse performance than SuperCon and do not demonstrate any performance gain when tested on the corrected labels. We speculate that, due to the relatively low mislabelling rates in these datasets (e.g.~5.85\% for Imagenet), these approaches may be overly heavy-handed in combating labelling noise, diminishing the models' performance as a result. D-SCL is comparatively more suitable for these lower noise rates and sees improved performance over these methods as a result.

\subsection{Transfer Learning}
\label{subsec:results-finetune}

We now assess performance when fine-tuning existing pre-trained models for specific downstream tasks.
Specifically, models are initialized with publicly-available weights from pretraining on ImageNet-21k \cite{DBLP:conf/cvpr/imagenet}, and are fine-tuned on smaller datasets using our objective.
We use 8 datasets: CIFAR-100 \citep{krizhevsky2009learning}, CUB-200-2011 \citep{WahCUB_200_2011}, Caltech-256 \citep{caltech256}, Oxford 102 Flowers \citep{DBLP:conf/icvgip/flowers}, Oxford-IIIT Pets \citep{DBLP:conf/cvpr/pets}, iNaturalist 2017 \citep{DBLP:conf/cvpr/inat2017}, Places365 \citep{DBLP:conf/nips/places}, and ImageNet-1k \citep{DBLP:conf/cvpr/imagenet}. We select BEiT-3 base \citep{beit3} as the image encoder due to its excellent performance on ImageNet-1k.  Similar to \cite{RN81}, our approach for fine-tuning pre-trained models with contrastive learning involves initially training the models using a contrastive learning loss, followed by training a linear layer atop the frozen trained models using cross-entropy loss.


Tab.~\ref{table:transfer} shows classification accuracies after fine-tuning with different methods. 
We see that D-SCL gives the best classification accuracy across all datasets, with particularly large improvements on iNat2017 (+3.4\%) and state-of-the-art performance on Places365 (+2.2\%) when compared to fine-tuning with SupCon \cite{RN81}.
Similar to the pre-training setting, the noise-robust objectives Sel-CL and TCL exhibit inferior performance compared to fine-tuning with cross-entropy or D-SCL across all 8 datasets.


\subsection{Robustness to Synthetic Noisy Labels}
\label{sec:robustness_noisy}

Whereas standard datasets exhibit only a moderate level of label noise (e.g.~5--10\%), we now examine performance at much higher noise rates ($>$18\%) than is typical in benchmarks.
As in Sec.~\ref{subsec:results-pretrain}, we use BEiT-3 trained from scratch without additional data.
Specifically, we use the artificially noisy datasets CIFAR-10N \cite{DBLP:journals/corr/abs-2110-12088} and CIFAR-100N \cite{DBLP:journals/corr/abs-2110-12088}.
Crucially, the label noise in these datasets is predominantly \textit{not} due to naturally-arising human errors, but rather to synthetic mislabeling.
This breaks our hypothesis (and foundational principle of D-SCL) that mislabelled samples usually have high visual similarity to their assigned class and occur naturally due to human errors.

Tab.~\ref{table:noisy dataset} shows that D-SCL outperforms cross-entropy and SuperCon on noisy and noise-free variants of these datasets. Again, we compare against Sel-CL and TCL and find that D-SCL's performance diminishes relative to these methods for high levels of label noise (e.g.~40\%). However, we show that D-SCL remains competitive with these methods at noise levels up to 18\% and significantly outperforms them when no noise is present. As such, we argue that D-SCL is more applicable and well suited to more realistic scenarios where the label noise rate is relatively low.


\begin{table}[t]
\centering
\resizebox{8.2cm}{!}{
\scriptsize
\begin{tabular}{@{}lccccc@{}}
\toprule
\multicolumn{1}{c}{\textbf{FT method}}&\multicolumn{3}{@{}c}{\textbf{CIFAR-10N}} &\multicolumn{2}{@{}c}{\textbf{CIFAR-100N}} \\
\cmidrule(r){2-4}\cmidrule(){5-6}
\multicolumn{1}{@{}r}{\textit{Noise level}} & \textit{Clean} & \textit{18\%}&\textit{40\%} & \textit{Clean} & \textit{40\%}  \\
\midrule
\multicolumn{1}{@{}c}{\textbf{CE}\cite{DBLP:journals/bstj/Shannon48}}&88.73&81.4& \multicolumn{1}{@{}l}{64.85}&60.24&40.80\\ 
\multicolumn{1}{@{}c}{\textbf{SupCon}\cite{RN81}}&94.14&86.31&\multicolumn{1}{@{}l}{66.49}&61.98&42.44\\ 
\multicolumn{1}{@{}c}{\textbf{Sel-CL}\cite{li2022selective}}&91.28&\textbf{90.49}&\multicolumn{1}{@{}l}{\textbf{87.16}}&60.5&57.52\\ 
\multicolumn{1}{@{}c}{\textbf{TCL}\cite{huang2023twin}}&90.48&89.72&\multicolumn{1}{@{}l}{86.33}&61.27&\textbf{57.89}\\ 
\multicolumn{1}{@{}c}{\textbf{D-SCL (ours)}}&\textbf{95.66}&88.75&\multicolumn{1}{@{}l}{68.92}&\textbf{69.41}&45.17\\ \bottomrule
\end{tabular}}
\caption{Performance of BEiT-3 base model when trained on datasets with different noise levels. All models here are trained from scratch using only the indicated dataset.}
\label{table:noisy dataset}
\end{table}

\section{Discussion and Conclusion}

\paragraph{Limitations.}
%
Although D-SCL exceeds the state-of-the-art, it still has certain limitations.
First, while existing research has estimated mislabeling rates for various datasets \cite{DBLP:conf/nips/NorthcuttAM21}, determining this for new datasets remains a challenge.
This issue could be effectively managed by adopting the typical average error rate of $3.3\%$, as reported in \cite{DBLP:conf/nips/NorthcuttAM21}, as a baseline for hyperparameter tuning to identify an optimal value. In our experiments, we observed that D-SCL's performance exhibits low sensitivity to the estimated mislabeling rates, with further details of this study provided in the supplementary materials.
Also, our method does not outperform the existing methods Sel-CL and TCL when applied to datasets with artificially inflated mislabeling rates.
This is because our model is tailored to mitigate labeling errors in datasets with noise characteristics typical in real-world scenarios---i.e.~where mislabelings occur naturally due to human error when classes are genuinely similar or ambiguous.

\paragraph{Conclusion.}
In this work, we investigated the extent and manner in which mislabelled samples impact supervised contrastive learning (SCL). Based on this, we introduced a novel, debiased objective, D-SCL, specifically designed to mitigate the influence of labeling errors and reduce sampling bias stemming from these errors. A significant advantage of D-SCL is its efficiency---it introduces no additional overhead during training. Our empirical results demonstrate that D-SCL not only excels as a supervised classification objective but also consistently outperform traditional cross-entropy methods and previous SCL approaches, both when training from scratch and when fine-tuning. Furthermore, D-SCL proves particularly beneficial for data-intensive models such as Vision Transformers (ViT) in scenarios with limited training data, showing its adaptability and effectiveness in challenging learning environments.



{
    \small
    \bibliographystyle{ieeenat_fullname}
    \bibliography{main}

\begin{thebibliography}{71}
\providecommand{\natexlab}[1]{#1}
\providecommand{\url}[1]{\texttt{#1}}
\expandafter\ifx\csname urlstyle\endcsname\relax
  \providecommand{\doi}[1]{doi: #1}\else
  \providecommand{\doi}{doi: \begingroup \urlstyle{rm}\Url}\fi

\bibitem[Adebayo et~al.(2023)Adebayo, Hall, Yu, and Chern]{DBLP:conf/iclr/AdebayoHYC23}
Julius Adebayo, Melissa Hall, Bowen Yu, and Bobbie Chern.
\newblock Quantifying and mitigating the impact of label errors on model disparity metrics.
\newblock In \emph{The Eleventh International Conference on Learning Representations, {ICLR} 2023, Kigali, Rwanda, May 1-5, 2023}. OpenReview.net, 2023.

\bibitem[Arazo et~al.(2019)Arazo, Ortego, Albert, O'Connor, and McGuinness]{DBLP:conf/icml/ArazoOAOM19}
Eric Arazo, Diego Ortego, Paul Albert, Noel~E. O'Connor, and Kevin McGuinness.
\newblock Unsupervised label noise modeling and loss correction.
\newblock In \emph{Proceedings of the 36th International Conference on Machine Learning, {ICML} 2019, 9-15 June 2019, Long Beach, California, {USA}}, pages 312--321. {PMLR}, 2019.

\bibitem[Arora et~al.(2019)Arora, Khandeparkar, Khodak, Plevrakis, and Saunshi]{DBLP:journals/corr/abs-1902-09229}
Sanjeev Arora, Hrishikesh Khandeparkar, Mikhail Khodak, Orestis Plevrakis, and Nikunj Saunshi.
\newblock A theoretical analysis of contrastive unsupervised representation learning.
\newblock \emph{CoRR}, abs/1902.09229, 2019.

\bibitem[Becker and Hinton(1992)]{becker1992self}
Suzanna Becker and Geoffrey~E Hinton.
\newblock Self-organizing neural network that discovers surfaces in random-dot stereograms.
\newblock \emph{Nature}, 355\penalty0 (6356):\penalty0 161--163, 1992.

\bibitem[Caron et~al.(2020)Caron, Misra, Mairal, Goyal, Bojanowski, and Joulin]{swav}
Mathilde Caron, Ishan Misra, Julien Mairal, Priya Goyal, Piotr Bojanowski, and Armand Joulin.
\newblock Unsupervised learning of visual features by contrasting cluster assignments.
\newblock In \emph{Advances in Neural Information Processing Systems 33: Annual Conference on Neural Information Processing Systems 2020, NeurIPS 2020, December 6-12, 2020, virtual}, 2020.

\bibitem[Chen et~al.(2021)Chen, Ye, Chen, Zhao, and Heng]{DBLP:conf/aaai/ChenYCZH21}
Pengfei Chen, Junjie Ye, Guangyong Chen, Jingwei Zhao, and Pheng{-}Ann Heng.
\newblock Beyond class-conditional assumption: {A} primary attempt to combat instance-dependent label noise.
\newblock In \emph{Thirty-Fifth {AAAI} Conference on Artificial Intelligence, {AAAI} 2021, Thirty-Third Conference on Innovative Applications of Artificial Intelligence, {IAAI} 2021, The Eleventh Symposium on Educational Advances in Artificial Intelligence, {EAAI} 2021, Virtual Event, February 2-9, 2021}, pages 11442--11450. {AAAI} Press, 2021.

\bibitem[Chen et~al.(2020{\natexlab{a}})Chen, Kornblith, Norouzi, and Hinton]{RN89}
Ting Chen, Simon Kornblith, Mohammad Norouzi, and Geoffrey~E. Hinton.
\newblock A simple framework for contrastive learning of visual representations.
\newblock In \emph{Proceedings of the 37th International Conference on Machine Learning, {ICML} 2020, 13-18 July 2020, Virtual Event}, pages 1597--1607. {PMLR}, 2020{\natexlab{a}}.

\bibitem[Chen et~al.(2020{\natexlab{b}})Chen, Fan, Girshick, and He]{mocov2}
Xinlei Chen, Haoqi Fan, Ross~B. Girshick, and Kaiming He.
\newblock Improved baselines with momentum contrastive learning.
\newblock \emph{CoRR}, abs/2003.04297, 2020{\natexlab{b}}.

\bibitem[Chuang et~al.(2020{\natexlab{a}})Chuang, Robinson, Lin, Torralba, and Jegelka]{DBLP:journals/corr/abs-2007-00224}
Ching{-}Yao Chuang, Joshua Robinson, Yen{-}Chen Lin, Antonio Torralba, and Stefanie Jegelka.
\newblock Debiased contrastive learning.
\newblock \emph{CoRR}, abs/2007.00224, 2020{\natexlab{a}}.

\bibitem[Chuang et~al.(2020{\natexlab{b}})Chuang, Robinson, Lin, Torralba, and Jegelka]{RN126}
Ching-Yao Chuang, Joshua Robinson, Yen-Chen Lin, Antonio Torralba, and Stefanie Jegelka.
\newblock Debiased contrastive learning.
\newblock \emph{Advances in neural information processing systems}, 33:\penalty0 8765--8775, 2020{\natexlab{b}}.

\bibitem[Deng et~al.(2009)Deng, Dong, Socher, Li, Li, and Fei{-}Fei]{DBLP:conf/cvpr/imagenet}
Jia Deng, Wei Dong, Richard Socher, Li{-}Jia Li, Kai Li, and Li Fei{-}Fei.
\newblock Imagenet: {A} large-scale hierarchical image database.
\newblock In \emph{2009 {IEEE} Computer Society Conference on Computer Vision and Pattern Recognition {(CVPR} 2009), 20-25 June 2009, Miami, Florida, {USA}}, pages 248--255. {IEEE} Computer Society, 2009.

\bibitem[Dosovitskiy et~al.(2021)Dosovitskiy, Beyer, Kolesnikov, Weissenborn, Zhai, Unterthiner, Dehghani, Minderer, Heigold, Gelly, Uszkoreit, and Houlsby]{rn83}
Alexey Dosovitskiy, Lucas Beyer, Alexander Kolesnikov, Dirk Weissenborn, Xiaohua Zhai, Thomas Unterthiner, Mostafa Dehghani, Matthias Minderer, Georg Heigold, Sylvain Gelly, Jakob Uszkoreit, and Neil Houlsby.
\newblock An image is worth 16x16 words: Transformers for image recognition at scale.
\newblock In \emph{9th International Conference on Learning Representations, {ICLR} 2021, Virtual Event, Austria, May 3-7, 2021}. OpenReview.net, 2021.

\bibitem[Du~Plessis et~al.(2014)Du~Plessis, Niu, and Sugiyama]{du2014analysis}
Marthinus~C Du~Plessis, Gang Niu, and Masashi Sugiyama.
\newblock Analysis of learning from positive and unlabeled data.
\newblock \emph{Advances in neural information processing systems}, 27, 2014.

\bibitem[Dubel et~al.(2023)Dubel, Wijata, and Nalepa]{DBLP:conf/iccS/DubelWN23}
Rafal Dubel, Agata~M. Wijata, and Jakub Nalepa.
\newblock On the impact of noisy labels on supervised classification models.
\newblock In \emph{Computational Science - {ICCS} 2023 - 23rd International Conference, Prague, Czech Republic, July 3-5, 2023, Proceedings, Part {II}}, pages 111--119. Springer, 2023.

\bibitem[Elkan and Noto(2008)]{elkan2008learning}
Charles Elkan and Keith Noto.
\newblock Learning classifiers from only positive and unlabeled data.
\newblock In \emph{Proceedings of the 14th ACM SIGKDD international conference on Knowledge discovery and data mining}, pages 213--220, 2008.

\bibitem[Ge et~al.(2018)Ge, Huang, Dong, and Scott]{DBLP:journals/corr/abs-1810-06951}
Weifeng Ge, Weilin Huang, Dengke Dong, and Matthew~R. Scott.
\newblock Deep metric learning with hierarchical triplet loss.
\newblock \emph{CoRR}, abs/1810.06951, 2018.

\bibitem[Griffin et~al.(2007)Griffin, Holub, and Perona]{caltech256}
Gregory Griffin, Alex Holub, and Pietro Perona.
\newblock \emph{Caltech-256 Object Category Dataset}.
\newblock 2007.

\bibitem[Grill et~al.(2020)Grill, Strub, Altch{\'{e}}, Tallec, Richemond, Buchatskaya, Doersch, Pires, Guo, Azar, Piot, Kavukcuoglu, Munos, and Valko]{byol}
Jean{-}Bastien Grill, Florian Strub, Florent Altch{\'{e}}, Corentin Tallec, Pierre~H. Richemond, Elena Buchatskaya, Carl Doersch, Bernardo~{\'{A}}vila Pires, Zhaohan Guo, Mohammad~Gheshlaghi Azar, Bilal Piot, Koray Kavukcuoglu, R{\'{e}}mi Munos, and Michal Valko.
\newblock Bootstrap your own latent - {A} new approach to self-supervised learning.
\newblock In \emph{Advances in Neural Information Processing Systems 33: Annual Conference on Neural Information Processing Systems 2020, NeurIPS 2020, December 6-12, 2020, virtual}, 2020.

\bibitem[Gunel et~al.(2020)Gunel, Du, Conneau, and Stoyanov]{suconlan}
Beliz Gunel, Jingfei Du, Alexis Conneau, and Ves Stoyanov.
\newblock Supervised contrastive learning for pre-trained language model fine-tuning.
\newblock \emph{CoRR}, abs/2011.01403, 2020.

\bibitem[Gutmann and Hyv{\"a}rinen(2010)]{gutmann2010noise}
Michael Gutmann and Aapo Hyv{\"a}rinen.
\newblock Noise-contrastive estimation: A new estimation principle for unnormalized statistical models.
\newblock In \emph{Proceedings of the thirteenth international conference on artificial intelligence and statistics}, pages 297--304. JMLR Workshop and Conference Proceedings, 2010.

\bibitem[Han et~al.(2020)Han, Yao, Liu, Niu, Tsang, Kwok, and Sugiyama]{DBLP:journals/corr/abs-2011-04406}
Bo Han, Quanming Yao, Tongliang Liu, Gang Niu, Ivor~W. Tsang, James~T. Kwok, and Masashi Sugiyama.
\newblock A survey of label-noise representation learning: Past, present and future.
\newblock \emph{CoRR}, abs/2011.04406, 2020.

\bibitem[He et~al.(2016)He, Zhang, Ren, and Sun]{RN36}
Kaiming He, Xiangyu Zhang, Shaoqing Ren, and Jian Sun.
\newblock Deep residual learning for image recognition.
\newblock In \emph{2016 {IEEE} Conference on Computer Vision and Pattern Recognition, {CVPR} 2016, Las Vegas, NV, USA, June 27-30, 2016}, pages 770--778. {IEEE} Computer Society, 2016.

\bibitem[Horn et~al.(2018)Horn, Aodha, Song, Cui, Sun, Shepard, Adam, Perona, and Belongie]{DBLP:conf/cvpr/inat2017}
Grant~Van Horn, Oisin~Mac Aodha, Yang Song, Yin Cui, Chen Sun, Alexander Shepard, Hartwig Adam, Pietro Perona, and Serge~J. Belongie.
\newblock The inaturalist species classification and detection dataset.
\newblock In \emph{2018 {IEEE} Conference on Computer Vision and Pattern Recognition, {CVPR} 2018, Salt Lake City, UT, USA, June 18-22, 2018}, pages 8769--8778. Computer Vision Foundation / {IEEE} Computer Society, 2018.

\bibitem[Houle(2017)]{DBLP:conf/sisap/Houle17}
Michael~E. Houle.
\newblock Local intrinsic dimensionality {I:} an extreme-value-theoretic foundation for similarity applications.
\newblock In \emph{Similarity Search and Applications - 10th International Conference, {SISAP} 2017, Munich, Germany, October 4-6, 2017, Proceedings}, pages 64--79. Springer, 2017.

\bibitem[Hu et~al.(2020)Hu, Li, and Yu]{RegularizeNoise}
Wei Hu, Zhiyuan Li, and Dingli Yu.
\newblock Simple and effective regularization methods for training on noisily labeled data with generalization guarantee.
\newblock In \emph{8th International Conference on Learning Representations, {ICLR} 2020, Addis Ababa, Ethiopia, April 26-30, 2020}. OpenReview.net, 2020.

\bibitem[Huang et~al.(2023)Huang, Zhang, and Shan]{huang2023twin}
Zhizhong Huang, Junping Zhang, and Hongming Shan.
\newblock Twin contrastive learning with noisy labels.
\newblock In \emph{Proceedings of the IEEE/CVF Conference on Computer Vision and Pattern Recognition}, pages 11661--11670, 2023.

\bibitem[Jenni and Favaro(2018{\natexlab{a}})]{Bilevel}
Simon Jenni and Paolo Favaro.
\newblock Deep bilevel learning.
\newblock In \emph{Computer Vision - {ECCV} 2018 - 15th European Conference, Munich, Germany, September 8-14, 2018, Proceedings, Part {X}}, pages 632--648. Springer, 2018{\natexlab{a}}.

\bibitem[Jenni and Favaro(2018{\natexlab{b}})]{DBLP:conf/eccv/JenniF18}
Simon Jenni and Paolo Favaro.
\newblock Deep bilevel learning.
\newblock In \emph{Computer Vision - {ECCV} 2018 - 15th European Conference, Munich, Germany, September 8-14, 2018, Proceedings, Part {X}}, pages 632--648. Springer, 2018{\natexlab{b}}.

\bibitem[Kalantidis et~al.(2020)Kalantidis, Sariyildiz, Pion, Weinzaepfel, and Larlus]{DBLP:journals/corr/abs-2010-01028}
Yannis Kalantidis, Mert~B{\"{u}}lent Sariyildiz, No{\'{e}} Pion, Philippe Weinzaepfel, and Diane Larlus.
\newblock Hard negative mixing for contrastive learning.
\newblock \emph{CoRR}, abs/2010.01028, 2020.

\bibitem[Khosla et~al.(2020)Khosla, Teterwak, Wang, Sarna, Tian, Isola, Maschinot, Liu, and Krishnan]{RN81}
Prannay Khosla, Piotr Teterwak, Chen Wang, Aaron Sarna, Yonglong Tian, Phillip Isola, Aaron Maschinot, Ce Liu, and Dilip Krishnan.
\newblock Supervised contrastive learning.
\newblock \emph{Advances in Neural Information Processing Systems}, 33:\penalty0 18661--18673, 2020.

\bibitem[Klie et~al.(2023)Klie, Webber, and Gurevych]{DBLP:journals/coling/KlieWG23}
Jan{-}Christoph Klie, Bonnie Webber, and Iryna Gurevych.
\newblock Annotation error detection: Analyzing the past and present for a more coherent future.
\newblock \emph{Comput. Linguistics}, 49\penalty0 (1):\penalty0 157--198, 2023.

\bibitem[Krizhevsky et~al.(2009)Krizhevsky, Hinton, et~al.]{krizhevsky2009learning}
Alex Krizhevsky, Geoffrey Hinton, et~al.
\newblock Learning multiple layers of features from tiny images.
\newblock 2009.

\bibitem[Li et~al.(2021)Li, Xiong, and Hoi]{MoPro}
Junnan Li, Caiming Xiong, and Steven C.~H. Hoi.
\newblock Mopro: Webly supervised learning with momentum prototypes.
\newblock In \emph{9th International Conference on Learning Representations, {ICLR} 2021, Virtual Event, Austria, May 3-7, 2021}. OpenReview.net, 2021.

\bibitem[Li et~al.(2022{\natexlab{a}})Li, Xia, Ge, and Liu]{SSCL}
Shikun Li, Xiaobo Xia, Shiming Ge, and Tongliang Liu.
\newblock Selective-supervised contrastive learning with noisy labels.
\newblock In \emph{{IEEE/CVF} Conference on Computer Vision and Pattern Recognition, {CVPR} 2022, New Orleans, LA, USA, June 18-24, 2022}, pages 316--325. {IEEE}, 2022{\natexlab{a}}.

\bibitem[Li et~al.(2022{\natexlab{b}})Li, Xia, Ge, and Liu]{li2022selective}
Shikun Li, Xiaobo Xia, Shiming Ge, and Tongliang Liu.
\newblock Selective-supervised contrastive learning with noisy labels.
\newblock In \emph{Proceedings of the IEEE/CVF Conference on Computer Vision and Pattern Recognition}, pages 316--325, 2022{\natexlab{b}}.

\bibitem[Lin et~al.(2023)Lin, Pi, Zhang, Xia, Gao, Zhou, Liu, and Han]{holviewtransmat}
Yong Lin, Renjie Pi, Weizhong Zhang, Xiaobo Xia, Jiahui Gao, Xiao Zhou, Tongliang Liu, and Bo Han.
\newblock A holistic view of label noise transition matrix in deep learning and beyond.
\newblock In \emph{The Eleventh International Conference on Learning Representations, {ICLR} 2023, Kigali, Rwanda, May 1-5, 2023}. OpenReview.net, 2023.

\bibitem[Liu et~al.(2021)Liu, Sangineto, Bi, Sebe, Lepri, and Nadai]{rn131}
Yahui Liu, Enver Sangineto, Wei Bi, Nicu Sebe, Bruno Lepri, and Marco Nadai.
\newblock Efficient training of visual transformers with small datasets.
\newblock \emph{Advances in Neural Information Processing Systems}, 34:\penalty0 23818--23830, 2021.

\bibitem[Liu et~al.(2023)Liu, Cheng, and Zhang]{transmat2}
Yang Liu, Hao Cheng, and Kun Zhang.
\newblock Identifiability of label noise transition matrix.
\newblock In \emph{International Conference on Machine Learning, {ICML} 2023, 23-29 July 2023, Honolulu, Hawaii, {USA}}, pages 21475--21496. {PMLR}, 2023.

\bibitem[Ma et~al.(2018)Ma, Wang, Houle, Zhou, Erfani, Xia, Wijewickrema, and Bailey]{DBLP:conf/icml/MaWHZEXWB18}
Xingjun Ma, Yisen Wang, Michael~E. Houle, Shuo Zhou, Sarah~M. Erfani, Shu{-}Tao Xia, Sudanthi N.~R. Wijewickrema, and James Bailey.
\newblock Dimensionality-driven learning with noisy labels.
\newblock In \emph{Proceedings of the 35th International Conference on Machine Learning, {ICML} 2018, Stockholmsm{\"{a}}ssan, Stockholm, Sweden, July 10-15, 2018}, pages 3361--3370. {PMLR}, 2018.

\bibitem[Nilsback and Zisserman(2008)]{DBLP:conf/icvgip/flowers}
Maria{-}Elena Nilsback and Andrew Zisserman.
\newblock Automated flower classification over a large number of classes.
\newblock In \emph{Sixth Indian Conference on Computer Vision, Graphics {\&} Image Processing, {ICVGIP} 2008, Bhubaneswar, India, 16-19 December 2008}, pages 722--729. {IEEE} Computer Society, 2008.

\bibitem[Northcutt et~al.(2021)Northcutt, Athalye, and Mueller]{DBLP:conf/nips/NorthcuttAM21}
Curtis~G. Northcutt, Anish Athalye, and Jonas Mueller.
\newblock Pervasive label errors in test sets destabilize machine learning benchmarks.
\newblock In \emph{Proceedings of the Neural Information Processing Systems Track on Datasets and Benchmarks 1, NeurIPS Datasets and Benchmarks 2021, December 2021, virtual}, 2021.

\bibitem[Oord et~al.(2018)Oord, Li, and Vinyals]{oord2018representation}
Aaron van~den Oord, Yazhe Li, and Oriol Vinyals.
\newblock Representation learning with contrastive predictive coding.
\newblock \emph{arXiv preprint arXiv:1807.03748}, 2018.

\bibitem[Oquab et~al.(2023)Oquab, Darcet, Moutakanni, Vo, Szafraniec, Khalidov, Fernandez, Haziza, Massa, El{-}Nouby, Assran, Ballas, Galuba, Howes, Huang, Li, Misra, Rabbat, Sharma, Synnaeve, Xu, J{\'{e}}gou, Mairal, Labatut, Joulin, and Bojanowski]{DBLP:journals/corr/abs-2304-07193}
Maxime Oquab, Timoth{\'{e}}e Darcet, Th{\'{e}}o Moutakanni, Huy Vo, Marc Szafraniec, Vasil Khalidov, Pierre Fernandez, Daniel Haziza, Francisco Massa, Alaaeldin El{-}Nouby, Mahmoud Assran, Nicolas Ballas, Wojciech Galuba, Russell Howes, Po{-}Yao Huang, Shang{-}Wen Li, Ishan Misra, Michael~G. Rabbat, Vasu Sharma, Gabriel Synnaeve, Hu Xu, Herv{\'{e}} J{\'{e}}gou, Julien Mairal, Patrick Labatut, Armand Joulin, and Piotr Bojanowski.
\newblock Dinov2: Learning robust visual features without supervision.
\newblock \emph{CoRR}, abs/2304.07193, 2023.

\bibitem[Ortego et~al.(2021)Ortego, Arazo, Albert, O'Connor, and McGuinness]{DBLP:conf/cvpr/OrtegoAAOM21}
Diego Ortego, Eric Arazo, Paul Albert, Noel~E. O'Connor, and Kevin McGuinness.
\newblock Multi-objective interpolation training for robustness to label noise.
\newblock In \emph{{IEEE} Conference on Computer Vision and Pattern Recognition, {CVPR} 2021, virtual, June 19-25, 2021}, pages 6606--6615. Computer Vision Foundation / {IEEE}, 2021.

\bibitem[Parkhi et~al.(2012)Parkhi, Vedaldi, Zisserman, and Jawahar]{DBLP:conf/cvpr/pets}
Omkar~M. Parkhi, Andrea Vedaldi, Andrew Zisserman, and C.~V. Jawahar.
\newblock Cats and dogs.
\newblock In \emph{2012 {IEEE} Conference on Computer Vision and Pattern Recognition, Providence, RI, USA, June 16-21, 2012}, pages 3498--3505. {IEEE} Computer Society, 2012.

\bibitem[Ren et~al.(2018)Ren, Zeng, Yang, and Urtasun]{DBLP:conf/icml/RenZYU18}
Mengye Ren, Wenyuan Zeng, Bin Yang, and Raquel Urtasun.
\newblock Learning to reweight examples for robust deep learning.
\newblock In \emph{Proceedings of the 35th International Conference on Machine Learning, {ICML} 2018, Stockholmsm{\"{a}}ssan, Stockholm, Sweden, July 10-15, 2018}, pages 4331--4340. {PMLR}, 2018.

\bibitem[Robinson et~al.(2021)Robinson, Chuang, Sra, and Jegelka]{DBLP:conf/iclr/RobinsonCSJ21}
Joshua~David Robinson, Ching{-}Yao Chuang, Suvrit Sra, and Stefanie Jegelka.
\newblock Contrastive learning with hard negative samples.
\newblock In \emph{9th International Conference on Learning Representations, {ICLR} 2021, Virtual Event, Austria, May 3-7, 2021}. OpenReview.net, 2021.

\bibitem[Russakovsky et~al.(2015)Russakovsky, Deng, Su, Krause, Satheesh, Ma, Huang, Karpathy, Khosla, Bernstein, Berg, and Li]{RN91}
Olga Russakovsky, Jia Deng, Hao Su, Jonathan Krause, Sanjeev Satheesh, Sean Ma, Zhiheng Huang, Andrej Karpathy, Aditya Khosla, Michael Bernstein, Alexander~C Berg, and Fei-Fei Li.
\newblock Imagenet large scale visual recognition challenge.
\newblock \emph{International journal of computer vision}, 115\penalty0 (3):\penalty0 211--252, 2015.

\bibitem[Schroff et~al.(2015)Schroff, Kalenichenko, and Philbin]{DBLP:journals/corr/SchroffKP15}
Florian Schroff, Dmitry Kalenichenko, and James Philbin.
\newblock Facenet: {A} unified embedding for face recognition and clustering.
\newblock \emph{CoRR}, abs/1503.03832, 2015.

\bibitem[Shannon(1948)]{DBLP:journals/bstj/Shannon48}
Claude~E. Shannon.
\newblock A mathematical theory of communication.
\newblock \emph{Bell Syst. Tech. J.}, 27\penalty0 (3):\penalty0 379--423, 1948.

\bibitem[Sohn(2016{\natexlab{a}})]{npairloss}
Kihyuk Sohn.
\newblock Improved deep metric learning with multi-class n-pair loss objective.
\newblock In \emph{Advances in Neural Information Processing Systems}. Curran Associates, Inc., 2016{\natexlab{a}}.

\bibitem[Sohn(2016{\natexlab{b}})]{sohn2016improved}
Kihyuk Sohn.
\newblock Improved deep metric learning with multi-class n-pair loss objective.
\newblock \emph{Advances in neural information processing systems}, 29, 2016{\natexlab{b}}.

\bibitem[Song et~al.(2023)Song, Kim, Park, Shin, and Lee]{DBLP:journals/tnn/SongKPSL23}
Hwanjun Song, Minseok Kim, Dongmin Park, Yooju Shin, and Jae{-}Gil Lee.
\newblock Learning from noisy labels with deep neural networks: {A} survey.
\newblock \emph{{IEEE} Trans. Neural Networks Learn. Syst.}, 34\penalty0 (11):\penalty0 8135--8153, 2023.

\bibitem[Song et~al.(2015)Song, Xiang, Jegelka, and Savarese]{DBLP:journals/corr/SongXJS15}
Hyun~Oh Song, Yu Xiang, Stefanie Jegelka, and Silvio Savarese.
\newblock Deep metric learning via lifted structured feature embedding.
\newblock \emph{CoRR}, abs/1511.06452, 2015.

\bibitem[Tan et~al.(2022)Tan, He, Bing, and Ng]{Tan2022DomainGF}
Qingyu Tan, Ruidan He, Lidong Bing, and Hwee~Tou Ng.
\newblock Domain generalization for text classification with memory-based supervised contrastive learning.
\newblock In \emph{International Conference on Computational Linguistics}, 2022.

\bibitem[Tanno et~al.(2019)Tanno, Saeedi, Sankaranarayanan, Alexander, and Silberman]{DBLP:conf/cvpr/TannoSSAS19}
Ryutaro Tanno, Ardavan Saeedi, Swami Sankaranarayanan, Daniel~C. Alexander, and Nathan Silberman.
\newblock Learning from noisy labels by regularized estimation of annotator confusion.
\newblock In \emph{{IEEE} Conference on Computer Vision and Pattern Recognition, {CVPR} 2019, Long Beach, CA, USA, June 16-20, 2019}, pages 11244--11253. Computer Vision Foundation / {IEEE}, 2019.

\bibitem[Wah et~al.(2011)Wah, Branson, Welinder, Perona, and Belongie]{WahCUB_200_2011}
C. Wah, S. Branson, P. Welinder, P. Perona, and S. Belongie.
\newblock {The Caltech-UCSD Birds-200-2011 Dataset}.
\newblock Technical Report CNS-TR-2011-001, California Institute of Technology, 2011.

\bibitem[Wang et~al.(2018)Wang, Liu, and Tao]{DBLP:journals/tnn/WangLT18}
Ruxin Wang, Tongliang Liu, and Dacheng Tao.
\newblock Multiclass learning with partially corrupted labels.
\newblock \emph{{IEEE} Trans. Neural Networks Learn. Syst.}, 29\penalty0 (6):\penalty0 2568--2580, 2018.

\bibitem[Wang et~al.(2022)Wang, Bao, Dong, Bjorck, Peng, Liu, Aggarwal, Mohammed, Singhal, Som, and Wei]{beit3}
Wenhui Wang, Hangbo Bao, Li Dong, Johan Bjorck, Zhiliang Peng, Qiang Liu, Kriti Aggarwal, Owais~Khan Mohammed, Saksham Singhal, Subhojit Som, and Furu Wei.
\newblock Image as a foreign language: Beit pretraining for all vision and vision-language tasks.
\newblock \emph{CoRR}, abs/2208.10442, 2022.

\bibitem[Wei et~al.(2021{\natexlab{a}})Wei, Tao, Xie, and An]{DBLP:conf/nips/WeiTXA21}
Hongxin Wei, Lue Tao, Renchunzi Xie, and Bo An.
\newblock Open-set label noise can improve robustness against inherent label noise.
\newblock In \emph{Advances in Neural Information Processing Systems 34: Annual Conference on Neural Information Processing Systems 2021, NeurIPS 2021, December 6-14, 2021, virtual}, pages 7978--7992, 2021{\natexlab{a}}.

\bibitem[Wei et~al.(2021{\natexlab{b}})Wei, Zhu, Cheng, Liu, Niu, and Liu]{DBLP:journals/corr/abs-2110-12088}
Jiaheng Wei, Zhaowei Zhu, Hao Cheng, Tongliang Liu, Gang Niu, and Yang Liu.
\newblock Learning with noisy labels revisited: {A} study using real-world human annotations.
\newblock \emph{CoRR}, abs/2110.12088, 2021{\natexlab{b}}.

\bibitem[Welinder and Perona(2010)]{welinder2010online}
Peter Welinder and Pietro Perona.
\newblock Online crowdsourcing: rating annotators and obtaining cost-effective labels.
\newblock In \emph{2010 IEEE Computer Society Conference on Computer Vision and Pattern Recognition-Workshops}, pages 25--32. IEEE, 2010.

\bibitem[Yao et~al.(2020)Yao, Liu, Han, Gong, Deng, Niu, and Sugiyama]{dual-t-transmatrix}
Yu Yao, Tongliang Liu, Bo Han, Mingming Gong, Jiankang Deng, Gang Niu, and Masashi Sugiyama.
\newblock Dual {T:} reducing estimation error for transition matrix in label-noise learning.
\newblock In \emph{Advances in Neural Information Processing Systems 33: Annual Conference on Neural Information Processing Systems 2020, NeurIPS 2020, December 6-12, 2020, virtual}, 2020.

\bibitem[Yao et~al.(2021)Yao, Sun, Zhang, Shen, Wu, Zhang, and Tang]{DBLP:conf/cvpr/YaoSZS00T21}
Yazhou Yao, Zeren Sun, Chuanyi Zhang, Fumin Shen, Qi Wu, Jian Zhang, and Zhenmin Tang.
\newblock Jo-src: {A} contrastive approach for combating noisy labels.
\newblock In \emph{{IEEE} Conference on Computer Vision and Pattern Recognition, {CVPR} 2021, virtual, June 19-25, 2021}, pages 5192--5201. Computer Vision Foundation / {IEEE}, 2021.

\bibitem[Yue and Jha(2022)]{DBLP:journals/corr/abs-2208-08464}
Chang Yue and Niraj~K. Jha.
\newblock {CTRL:} clustering training losses for label error detection.
\newblock \emph{CoRR}, abs/2208.08464, 2022.

\bibitem[Zbontar et~al.(2021)Zbontar, Jing, Misra, LeCun, and Deny]{barlowtwins}
Jure Zbontar, Li Jing, Ishan Misra, Yann LeCun, and St{\'{e}}phane Deny.
\newblock Barlow twins: Self-supervised learning via redundancy reduction.
\newblock In \emph{Proceedings of the 38th International Conference on Machine Learning, {ICML} 2021, 18-24 July 2021, Virtual Event}, pages 12310--12320. {PMLR}, 2021.

\bibitem[Zhang et~al.(2017{\natexlab{a}})Zhang, Bengio, Hardt, Recht, and Vinyals]{rethinkgeneralization}
Chiyuan Zhang, Samy Bengio, Moritz Hardt, Benjamin Recht, and Oriol Vinyals.
\newblock Understanding deep learning requires rethinking generalization.
\newblock In \emph{5th International Conference on Learning Representations, {ICLR} 2017, Toulon, France, April 24-26, 2017, Conference Track Proceedings}. OpenReview.net, 2017{\natexlab{a}}.

\bibitem[Zhang et~al.(2017{\natexlab{b}})Zhang, Ciss{\'{e}}, Dauphin, and Lopez{-}Paz]{DBLP:journals/corr/abs-1710-09412}
Hongyi Zhang, Moustapha Ciss{\'{e}}, Yann~N. Dauphin, and David Lopez{-}Paz.
\newblock mixup: Beyond empirical risk minimization.
\newblock \emph{CoRR}, abs/1710.09412, 2017{\natexlab{b}}.

\bibitem[Zheng et~al.(2020)Zheng, Wu, Goswami, Goswami, Metaxas, and Chen]{DBLP:conf/icml/ZhengWG0MC20}
Songzhu Zheng, Pengxiang Wu, Aman Goswami, Mayank Goswami, Dimitris~N. Metaxas, and Chao Chen.
\newblock Error-bounded correction of noisy labels.
\newblock In \emph{Proceedings of the 37th International Conference on Machine Learning, {ICML} 2020, 13-18 July 2020, Virtual Event}, pages 11447--11457. {PMLR}, 2020.

\bibitem[Zhou et~al.(2014)Zhou, Lapedriza, Xiao, Torralba, and Oliva]{DBLP:conf/nips/places}
Bolei Zhou, {\`{A}}gata Lapedriza, Jianxiong Xiao, Antonio Torralba, and Aude Oliva.
\newblock Learning deep features for scene recognition using places database.
\newblock In \emph{Advances in Neural Information Processing Systems 27: Annual Conference on Neural Information Processing Systems 2014, December 8-13 2014, Montreal, Quebec, Canada}, pages 487--495, 2014.

\bibitem[Zhu et~al.(2023)Zhu, Chen, and Yang]{rn130}
Haoran Zhu, Boyuan Chen, and Carter Yang.
\newblock Understanding why vit trains badly on small datasets: An intuitive perspective.
\newblock \emph{arXiv preprint arXiv:2302.03751}, 2023.

\end{thebibliography}
}

\clearpage
\setcounter{page}{1}
\maketitlesupplementary


\begin{figure*}[t]
  \centering
  \includegraphics[width=\linewidth]{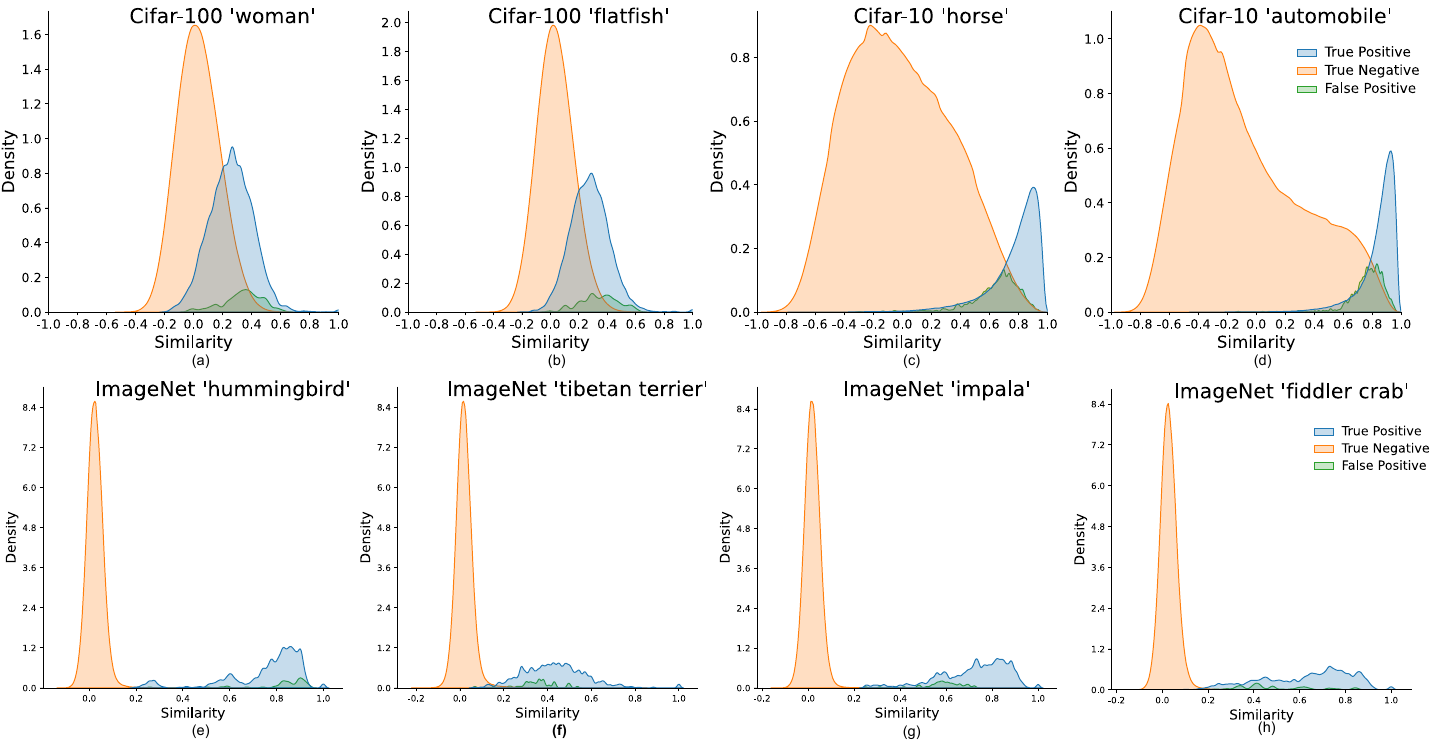}
  \caption{Distributions of similarities for true positive pairs, true negative pairs and false positives (mislabeled samples paired with samples from the assigned label class), for CIFAR-10, CIFAR-100 and ImageNet. The overlap is high between true positives and false positives.}
  \label{fig:similaritymap_imagenet}
\end{figure*}

\section{Supplement to Sec.~4: Analysing Label Errors and their Impacts }

This section extends our exploration in Sec.~4 of main paper, focusing on the similarities between different types of  pairs, including true positives, true negatives and false positives (mislabeled samples paired with samples from the assigned label class). We perform both a qualitative and a quantitative assessment.
For qualitative assessment, we plot  distributions of similarities between pairs, to intuitively show the relationships among different kinds of pairs.
For quantitative assessment, we use the Jensen-Shannon Divergence (JSD) to precisely measure these similarities.
Combined, these provide a comprehensive insight into the relationship between false positives and true positives.
Overall this analysis validates our hypothesis: false positive pairs are very similar to true positive pairs, thus, shown as ``easy positive" , in relation to the anchor, owing to their strong visual resemblance.

\subsection{Qualitative Assessment for Similarity Between Different Pairs}

In the main paper, we hypothesize that false positive samples are grouped with the positive set in relation to the anchor due to their similar visual characteristics, making accurate labelling difficult for both humans and machines.
Fig.~\ref{fig:similaritymap_imagenet} (a)--(d) show the similarity maps for the CIFAR-100 and CIFAR-10 datasets.
Fig.~\ref{fig:similaritymap_imagenet} (e)--(h) shows similarity maps for the more diverse and fine-grained ImageNet dataset.
Each figure is a plot of the distribution of the cosine similarities between pairs of different kinds -- true positive pairs, true negative pairs, and false positive pairs (mislabeled samples paired with samples from the assigned label class). On all three datasets, we see that false positive samples have a much greater similarity overlap with true positive pairs than with negative ones. This observation reinforces our hypothesis qualitatively. Furthermore, this insight guides the design of our method, which incorporates less weighting on easy positives and reduces bias caused by labeling errors.

\subsection{Quantitative Assessment for Similarity Between Different Pairs}

Next we measure the association between different types of pairs quantitatively, using Jensen-Shannon divergence (JSD). JSD is a widely-recognized measure for quantifying the similarity between probability distributions and a higher value means a larger distance. In Tab~\ref{tab:JSD}, we present the average JSD values among all the classes in each dataset for comparing similarities of true positive pairs with those of true negative pairs and false positive pairs, on CIFAR-10, CIFAR-100 and ImageNet.
Notably, the divergence between true positives and false positives pairs (0.29) is much smaller compared to true positives and true negatives (0.74) on ImageNet dataset. 
These findings quantitatively strengthen our hypothesis, leading us to implement the 'debiasing positive' approach, which can mitigate the negative impact of mislabeled samples, thereby enhancing the precision and reliability of our classification models.


\begin{table}[t]
\centering
\resizebox{\linewidth}{!}{%
{\tiny 
\begin{tabular}{@{}lccccc@{}}
\toprule
\textbf{Dataset} & \textbf{Pos \& Neg} & \textbf{Pos \& False Pos}\\
\midrule
CIFAR-10  & 0.56 & 0.24  \\ 
CIFAR-100 & 0.41 & 0.10 \\
ImageNet  & 0.74 & 0.29 \\
\bottomrule
\end{tabular}
}
}
\caption{the average JSD values comparing True Positives to True Negatives and True Positives to False Positives across all classes in each dataset. `Pos' stands for true positive samples, and `Neg' stands for true negative samples, while `False Pos' refers to False positive pairs (Mislabeled samples).}
\label{tab:JSD}
\end{table}

\section{Supplement to Sec.~5: Debiased Contrastive Learning}

\subsection{Contrastive Learning Setup}

\begin{align}
    q^{+} (x^{+}) := \,& q(x^{+}\,|\,z(x) = z(x^{+})) \\
    \propto \,& \frac{1}{e^{\beta f(x)^{T}f(x^{+})}} \cdot p(x^{+})
    \label{eq:q}
\end{align}

The exponential term in Eq.\ref{eq:q} is an unnormalized von Mises-Fisher distribution with a mean direction of \( f(x) \) and a concentration parameter \( \beta \). 
    This term increases the probability of sampling hard positives, (\textbf{P2}).
    The concentration parameter \( \beta \) modulates the weighting scheme of \( q^{+} \), specifically augmenting the weights of instances \( x^{+} \) that exhibit a lower inner product (i.e.~greater dissimilarity) with the anchor point \( x \). Given that the function \( f \) resides on the surface of a hypersphere with radius \( 1/t \), the squared Euclidean distance between \( f(x) \) and \( f(x') \) can be expressed as \( \| f(x) - f(x') \|^{2} = 2/t^{2} - 2f(x)^{T}f(x') \). The scaling hyperparameter \( t \) controls the radius of the resulting hypersphere. Thus, prioritizing instances with lower inner products is mathematically equivalent to favoring those with greater squared Euclidean distances (i.e.~greater dissimilarity).

\subsection{Contrastive Learning Principles}

In this section, we describe our approach to reduce sampling bias caused by noisy labels and how this fits into our overall debiased contrastive learning objective. Our proposed objective is rooted in four principles, of which the first three have been empirically supported by numerous studies~\cite{RN89,RN91,DBLP:journals/corr/abs-2007-00224,DBLP:journals/corr/abs-2010-01028}, while the fourth is introduced in our paper (and verified in Sec.~4.2). 
and goes beyond prior works:
\begin{enumerate}

    \item \textbf{True Positives with Latent Class Focus}: It is imperative that positive samples belong to the same latent class as the anchor \(x\).
    \item \textbf{True Negatives with Latent Class Focus}: It is important that negative samples belong to a different latent class than the anchor \(x\). 
     \item \textbf{Focus on Challenging Negative Samples}: The core of effective contrastive learning is the model's ability to discern between closely related samples. Therefore, negative samples that the model's current embedding perceives as akin to the anchor, termed "hard" negative samples, are the most instructive. They push the model's boundaries, facilitating more refined feature extraction. 
     \item \textbf{Deprioritizing Easy Positives}: The positive samples that the model's embedding perceives as akin to the anchor, are treated as ``easy" positive samples. By reducing the weighting of the easy positive samples, we minimize the effect of wrong learning signals from false positive pairs. Moreover, this approach compels the model to recognize and encode deeper, inherent similarities that are not immediately apparent, thereby enhancing its discernment capabilities.
\end{enumerate}

Our proposed objective, D-SCL achieves the outlined principles by striving to ensure true positive and true negative samples through debiasing the labeling error. It deprioritizes easy positives to minimize incorrect learning signals and prioritizes hard negatives to enhance its discernment capabilities. This approach fosters a debiased and effective contrastive learning process.

\subsection{Debiasing negatives}

We now describe how to debias the contrastive loss for negative samples, and simultaneously mine hard negatives.
We take an analogous approach to that for positives (which was described in the main paper).
We begin with Eq.~\ref{eq:supercon_weighted} explained in the main paper. This equation, termed the \textit{true label loss}, represents the ideal loss function which is grounded in the true prior distribution, $p$:

\begin{equation}
\resizebox{0.45\textwidth}{!}{$
\mathcal{L}_{T}  = \mathbb{E}_{\substack{x\sim p \\ x_{k}^{+} \sim  p_{x}^{+} \\ x_{i}^{-} \sim  p_{x}^{-}}}   \left[   \frac{-1}{\left |K  \right |}       \log \frac{\frac{Q}{K}  \sum_{k=1}^{K}     e^{f(x)^T f(x_{k}^{+})}}{\frac{Q}{K} \sum_{k=1}^{K} e^{f(x)^T f( x_{k}^{+})}+ \frac{W}{N}\sum_{i=1}^N e^{f(x)^T f(x_i^{-})}}\right]
$}
\label{eq:supercon_weighted} 
\end{equation}
Holding $W$ fixed and taking $N$ to be sufficiently large, we get
\begin{equation}
\resizebox{0.45\textwidth}{!}{$
\mathcal{L}_{T}  = \mathbb{E}_{\substack{x\sim p \\ x_{k}^{+} \sim  p_{x}^{+} \\ x_{i}^{-} \sim  p_{x}^{-}}}   \left[   \frac{-1}{\left |K  \right |}       \log \frac{\frac{Q}{K}  \sum_{k=1}^{K}     e^{f(x)^T f(x_{k}^{+})}}{\frac{Q}{K} \sum_{k=1}^{K} e^{f(x)^T f( x_{k}^{+})}+ W\sum_{i=1}^N e^{f(x)^T f(x_i^{-})}}\right]
$}
\label{eq:negative_sampling} 
\end{equation}



We first define \( q^{+}\left(x^{-}\right)=q\left(x^{-} \mid h(x)=h\left(x^{-}\right)\right) \propto e^{\beta f(x)^{\top} f\left(x^{-}\right)} \cdot p^{+}\left(x^{-}\right) \).
Then, by conditioning on the event \( \{ z(x) = z(x^{-}) \} \), we can write 

\begin{align}
    q\left(x^{-}\right) &= \tau^{-} q^{-}\left(x^{-}\right)+\tau^{+} q^{+}\left(x^{-}\right)
    \label{eq:neg_split}
\\ \Rightarrow \;
q^{-}\left(x^{-}\right)&=\left(q\left(x^{-}\right)-\tau^{+} q^{+}\left(x^{-}\right)\right) / \tau^{-}
 \label{eq:neg_rerrange}
\end{align}
By substituting Eq.~\ref{eq:neg_rerrange} into Eq.~\ref{eq:negative_sampling}, we obtain an objective that removes bias from labeling errors and also down-weights easy negative: 
\begin{equation}\
\resizebox{0.43\textwidth}{!}{$
\mathbb{E}_{\substack{x\sim p\\ x^{+} \sim  q \\ x^{-} \sim  q}}    \left[  \frac{-1}{\left |K  \right |}       \log \frac{\sum_{k=1}^{K}     e^{f(x)^T f\left(x_{k}^{+}\right)}}{\sum_{k=1}^{K} e^{f(x)^T f\left( x_{k}^{+}\right)}        +    \frac{W}{\tau^{-}}\left(\mathbb{E}_{x^{-} \sim q}\left[e^{f(x)^T f\left(x^{-}\right)}\right]-\tau^{+} \mathbb{E}_{b \sim q^{+}}\left[e^{f(x)^T f(b)}\right]\right) }   \right]
    $}
    \label{eq:hard_neg}
\end{equation}
To approximate the expectations $\mathbb{E}_{x^{-} \sim q}\left[e^{f(x)^T f\left(x^{-}\right)}\right]$ and $\mathbb{E}_{b \sim q^{+}}\left[e^{f(x)^T f(b)}\right]$ over $q$ and $q^{+}$, which can be achieved by classical Monte Carlo importance sampling, using samples from \(p\) and \(p^{+}\)  through Eq.~\ref{eq:negative_monte1} and \ref{eq:negative_monte2}:
\begin{equation}
\resizebox{0.43\textwidth}{!}{$
    \mathbb{E}_{x^{-} \sim q}\left[e^{f(x)^T f\left(x^{-}\right)}\right]=\mathbb{E}_{x^{-} \sim p}\left[e^{f(x)^T f\left(x^{-}\right)} q / p\right]=\mathbb{E}_{x^{-} \sim p}\left[e^{(\beta+1) f(x)^T f\left(x^{-}\right)} / Z\right]
        $}
        \label{eq:negative_monte1}
\end{equation}
\begin{equation}
\resizebox{0.43\textwidth}{!}{$
\mathbb{E}_{b \sim q^{+}}\left[e^{f(x)^T f(b)}\right]=\mathbb{E}_{b \sim p^{+}}\left[e^{f(x)^T f(b)} q^{+} / p^{+}\right]=\mathbb{E}_{b \sim p^{+}}\left[e^{(\beta+1) f(x)^T f(b)} / Z^{+}\right]
        $}
        \label{eq:negative_monte2}
\end{equation}
The remaining unknowns, namely the partition functions $Z=\mathbb{E}_{x^{-} \sim p}\left[e^{\beta f(x)^T f\left(x^{-}\right)}\right]$ and $Z^{+}=\mathbb{E}_{x^{+} \sim p^{+}}\left[e^{\beta f(x)^T f\left(x^{+}\right)}\right]$ admit empirical estimates:
\begin{align}
\widehat{Z}(x) &=\frac{1}{N} \sum_{i=1}^N e^{\beta f(x)^{\top} f\left(x_i^{-}\right)}
    \\
\widehat{Z}^{+}(x) &=\frac{1}{M} \sum_{i=1}^M e^{\beta f(x)^{\top} f\left(x_i^{+}\right)}
\end{align}
\begin{figure*}
\small
\centering
\begin{equation}
\mathbb{E}_{\substack{x\sim p\\ x^{+} \sim  q \\ x^{-} \sim  q}} \left[ \log \frac{-1}{\left |K  \right |}  \frac{\frac{Q}{\tau^{+}}\left(\mathbb{E}_{x^{+} \sim q}\left[e^{f(x)^T f(x^{+})}\right] - \tau^{-} \mathbb{E}_{v \sim q^{-}}\left[e^{f(x)^T f(v)}\right]\right)}{\frac{Q}{\tau^{+}}\left(\mathbb{E}_{x^{+} \sim q}\left[e^{f(x)^T f(x^{+})}\right] - \tau^{-} \mathbb{E}_{v \sim q^{-}}\left[e^{f(x)^T f(v)}\right]\right) + \frac{W}{\tau^{-}}\left(\mathbb{E}_{x^{-} \sim q}\left[e^{f(x)^T f(x^{+})}\right] - \tau^{+} \mathbb{E}_{b \sim q^{+} }\left[e^{f(x)^T f(b)}\right]\right)}\right]
\label{eq:easy_contrastivelearning}
\end{equation}
\end{figure*}

\paragraph{Overall learning objective.}
Using Eq.~11 (from the main paper) and Eq.~\ref{eq:hard_neg}, we get the final debiased contrastive learning objective Eq.~\ref{eq:easy_contrastivelearning}.

\section{Additional Experiments}

\subsection{Ablation Study}

\begin{table}[t]
\centering
\resizebox{\linewidth}{!}{
\begin{tabular}{@{}cccc|cc@{}}
\toprule
 \textbf{CE} & \textbf{SCL} & \textbf{\begin{tabular}[c]{@{}c@{}}Debiasing Pos\end{tabular}} & \textbf{\begin{tabular}[c]{@{}c@{}}Debiasing Neg \end{tabular}} & \textbf{CIFAR-100} & \textbf{ImageNet-1k} \\ \midrule
 \checkmark   & -            & -                                                                        & -                                                                        & 59.67             & 77.91                \\
 -           & \checkmark    & -                                                                        & -                                                                        & 60.25             & 81.73                \\
 -           & \checkmark    & -                                                                        & \checkmark                                                                & 61.30              & 83.25                \\
 -           & \checkmark    & \checkmark                                                                & -                                                                        & 63.18             & 83.61                \\
-           & \checkmark    & \checkmark                                                                & \checkmark                                                                & 64.47             & 84.21                \\ \bottomrule
\end{tabular}}
\caption{Results on CIFAR-100 and ImageNet-1k when training BEiT-3 base model from scratch using ablated versions of our D-SCL. SCL with no positive or negative mining refers to the loss proposed by \cite{RN81}. }
\label{tab:ablation}
\end{table}

We conduct an ablation study on CIFAR-100 and ImageNet-1k to measure the benefit of two key parts of our approach---debiasing in positives and debiasing in negatives (Tab.~\ref{tab:ablation}). When removing debiasing of both positives and negatives, we recover the original loss $L\_in$ proposed by \cite{RN81}. For both datasets, we see that debiasing in positives plays a more important role than debiasing in negatives---e.g.~on CIFAR-100, accuracy is improved from 60.3\% to 63.2\% with debiasing in positives but only to 61.3\% with debiasing in negatives. Nonetheless, we do find that each leads to a performance increase and that combining both, as in our full D-SCL, leads to the best performance. The relative importance of debiasing in positives over negatives validates our hypothesis that mislabelled samples impact the performance of supervised contrastive learning due to their occurrence as soft positives.

\subsection{Impact of Batch Size}
We conduct a comprehensive analysis to evaluate the influence of batch sizes on the transfer learning performance of the cross-entropy (CE) and our novel D-SCL objective across the CIFAR-100 and iNat2017 datasets. The findings are detailed in Table~\ref{tab:batch_size}.

Our analysis indicates that while there is a slight increase in the accuracy of D-SCL with larger batch sizes on both datasets, this improvement is modest, suggesting that the batch size does not severely influence the performance of our proposed objective. This implies that D-SCL maintains robustness even with relatively small batch sizes, such as 64, which is feasible for a GPU with 24 GB of memory. Notably, when comparing the models at a constant batch size of 64, D-SCL consistently surpassed CE in accuracy on both datasets. This underscores the effectiveness of our D-SCL objective, affirming its utility even with constrained batch sizes.

\begin{table}[t]
\centering
\resizebox{0.85\linewidth}{!}{%
{
\begin{tabular}{@{}l|ccc@{}}
\toprule
\textbf{FT method} & \textbf{Batch Size}
 & \textbf{CIFAR-100} & \textbf{INat2017} \\ \midrule
CE\cite{DBLP:journals/bstj/Shannon48}                 & 64              & 92.96            & 72.31  \\
 D-SCL (ours)                 & 32              & 92.71              & 72.74  \\
  D-SCL (ours)                 &64              & 93.62              & 74.90  \\
 D-SCL (ours)          & 128              & 93.81            & 76.25              \\
 \bottomrule
\end{tabular}}}
\caption{Classification accuracy of BEiT-3 model with different batch size, on CIFAR-100 and iNat2017.}
\label{tab:batch_size}
\end{table}

\subsection{Impact of imprecise estimated mislabeling rate}

\begin{table}[t]
\centering
\resizebox{0.7\linewidth}{!}{%
{
\begin{tabular}{@{}l|ccc@{}}
\toprule
\textbf{FT method} & $\boldsymbol{\tau^{\scalebox{0.5}{$+$}}}$
 & \textbf{CIFAR-10} & \textbf{CIFAR-100} \\ \midrule
CE\cite{DBLP:journals/bstj/Shannon48}                 & N/A              & 71.70            & 59.67  \\
 D-SCL (ours)                 & 0.01              & \textbf{90.16 }              & 63.92  \\
  D-SCL (ours)                 & \textbf{0.03}              & 90.03              & 64.16  \\
 D-SCL (ours)          & 0.05              & 89.29            & \textbf{64.47}              \\
 D-SCL (ours)         & 0.10              & 87.73            & 64.10       \\
 D-SCL (ours)          & 0.15              & 87.41            & 62.76   \\
 D-SCL (ours)           & 0.20              & 86.11            & 61.38    \\ \bottomrule
\end{tabular}}}
\caption{Classification accuracy of BEiT-3 model with different value of $\tau^+$, on CIFAR-10 and CIFAR-100.}
\label{tab:tau}
\end{table}

As we pointed out in the Sec.~7 in the main paper, determining the mislabeling rate for new datasets remains a challenge. Therefore, in this section, we measure the robustness of our method to inaccurate estimates of this parameter.
Based on \cite{DBLP:conf/nips/NorthcuttAM21}, we assume a true mislabeling rate of 3.3\%.
We vary the assumed mislabelling rate $\tau^{+}$ above and below this, and measure the impact this has on classification accuracy for CIFAR-10 and CIFAR-100.

Results are presented in Tab.~\ref{tab:tau}.
We explore a range of $\tau^{+}$ values from 0.01 to 0.20. We see $\tau^{+} = 0.01$ yields the highest accuracy for D-SCL on CIFAR-10, while $\tau^{+}=0.05$ was most effective for CIFAR-100. Notably, when $\tau^{+}$ is set to 0.03, which aligns with the average error rate, the classification accuracy on both CIFAR-10 and CIFAR-100 is very slightly lower compared to the best performing settings---0.13\% lower for CIFAR-10 at $\tau^{+} = 0.01$ and 0.31\% lower for CIFAR-100 at $\tau^{+}=  0.05$. We draw two conclusions.
Firstly, D-SCL shows consistent performance across different mislabeling rates, demonstrating a low sensitivity to these variations. This underscores the robustness of D-SCL in the face of potential mislabeling errors.
Secondly, our results indicate that the specific choice of $\tau^{+}$ is not critical to the overall success of the model. Even when $\tau^{+}$ is not set to its optimal value, D-SCL's performance still greatly exceeds that achieved using the Cross-Entropy (CE) loss function. This suggests that while tuning $\tau^{+}$ for a given dataset may be advantageous, the general efficacy of D-SCL is not heavily reliant on finding the exact value of $\tau^{+}$.
Thus, D-SCL may be straightforwardly used in practical scenarios where the mislabeling rate is uncertain.


\end{document}